\documentclass[11pt]{article}

\usepackage[final]{acl}

\usepackage{times}
\usepackage{latexsym}

\usepackage[T1]{fontenc}

\usepackage[utf8]{inputenc}

\usepackage{microtype}

\usepackage{inconsolata}

\usepackage{graphicx}
\usepackage{amsmath}
\usepackage{multirow}
\usepackage{booktabs}
\usepackage{enumitem}
\usepackage[table]{xcolor}
\usepackage[most]{tcolorbox}
\usepackage{CJKutf8}
\usepackage{tabularx}
\usepackage{makecell}

\definecolor{highpartial}{RGB}{210, 105, 30} 
\definecolor{lowpartial}{RGB}{220, 20, 60}   
\definecolor{fullmatch}{RGB}{0, 128, 128}    

\definecolor{mycolor}{gray}{0.85} 

\newcommand{\bb}[1]{\textbf{#1}} 
\newcommand{\cc}[1]{{\setlength{\fboxsep}{1pt}\colorbox{mycolor}{#1}}}
\newcommand{\bbcc}[1]{{\setlength{\fboxsep}{1pt}\colorbox{mycolor}{\textbf{#1}}}}
\newcommand{\cn}[1]{\begin{CJK*}{UTF8}{gbsn}#1\end{CJK*}}

\title{M\textsuperscript{2}PO: Multi-Perspective Multi-Pair Preference Optimization \\ for Machine Translation}

\author{
  \textbf{Hao Wang}$^{1}$, \textbf{Linlong Xu}$^{1}$, \textbf{Heng Liu}$^{1}$, \textbf{Yangyang Liu}$^{1}$, \textbf{Xiaohu Zhao}$^{1}$, \\
  \textbf{Bo Zeng}$^{1}$, \textbf{Liangying Shao}$^{1}$, \textbf{Yichen Dong}$^{1}$, \textbf{Xinwei Wu}$^{1}$, \textbf{Jiang Zhou}$^{1}$, \\
  \textbf{Tianyu Dong}$^{1}$, \textbf{Xiangxiang Zeng}$^{2}$, \textbf{Longyue Wang}$^{1}$, \textbf{Weihua Luo}$^{1}$ \\
  $^{1}$Alibaba Group, $^{2}$Hunan University \\
  \texttt{huaiyu.wh@alibaba-inc.com}
}

\begin{document}
\maketitle

\begin{abstract}
Aligning Large Language Models (LLMs) to human preferences is pivotal for Machine Translation (MT), yet current approaches are often hindered by misleading reward signals. Our analysis reveals that prevailing Quality Estimation (QE) models exhibit a systematic blind spot towards \textbf{partial errors}—specifically partial hallucinations and omissions—often favoring superficially fluent but unfaithful translations. To address this, we propose \textbf{M\textsuperscript{2}PO} (\textbf{M}ulti-Perspective \textbf{M}ulti-Pair \textbf{P}reference \textbf{O}ptimization), a data-centric framework for preference optimization in machine translation. First, to correct the bias towards fluency, M\textsuperscript{2}PO uses a multi-perspective alignment mechanism that decouples semantic fidelity from fluency, prioritizing faithfulness via a curriculum strategy. Second, with the bias corrected, partial errors fall between perfect and severely incorrect translations, making them inefficient to learn via standard best-versus-worst comparisons. We thus introduce a multi-pair objective that leverages the full candidate list to capture these fine-grained error signals. Experiments on WMT23, WMT24, and FLORES-200 show that M\textsuperscript{2}PO enables a 9B model to outperform leading open-source baselines and achieve parity with proprietary models like GPT-4o and Gemini-2.0-Flash, demonstrating significant potential for efficient, high-fidelity LLM-based translation\footnote{Our code is available at \url{https://github.com/AIDC-AI/Marco-MT/tree/master/MMPO}.}.
\end{abstract}

\section{Introduction}

\begin{figure}[t]
    \centering
    \includegraphics[width=1\columnwidth]{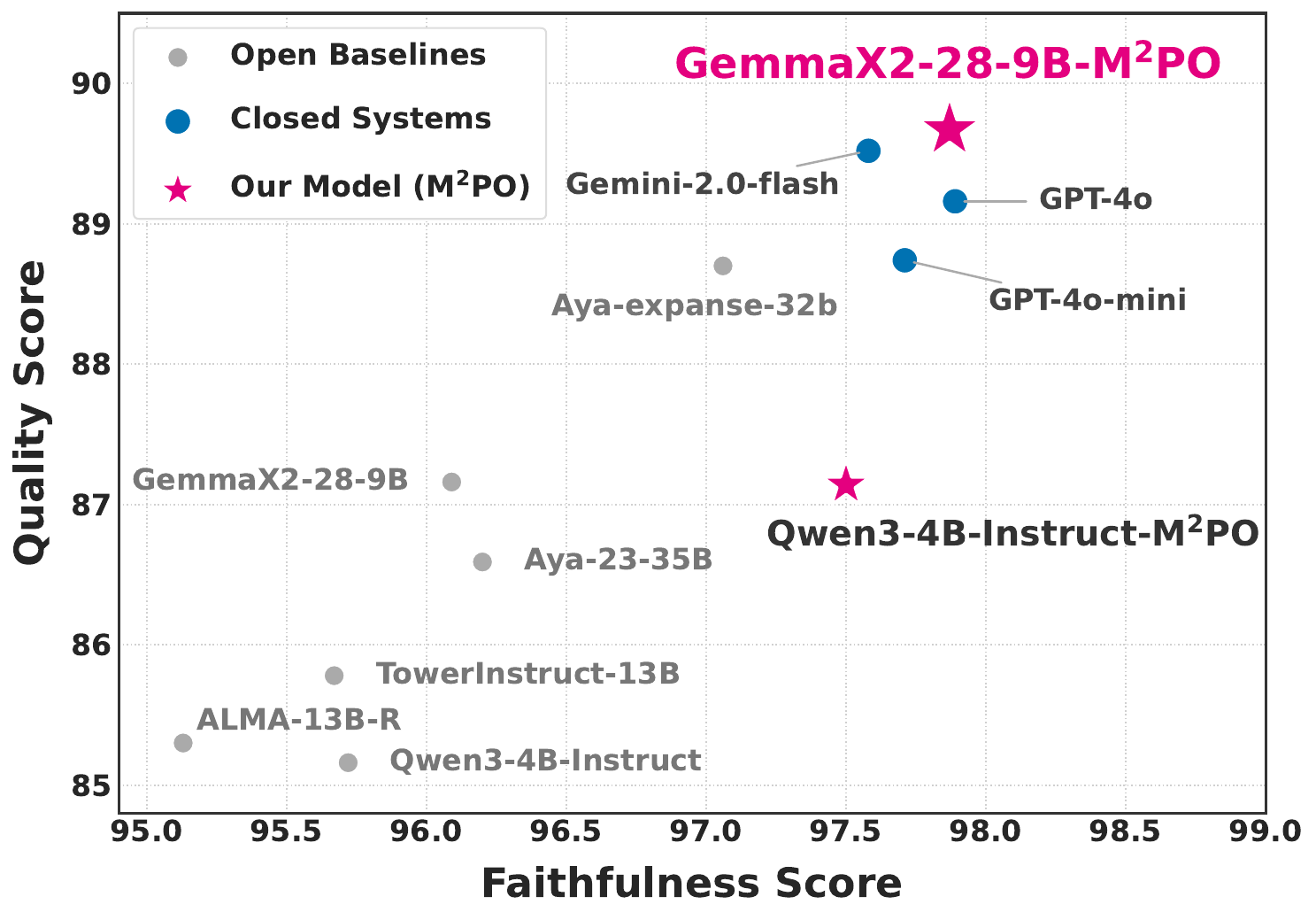}
    \caption{Translation Quality (XCOMET) vs. Faithfulness (Coverage Score~\citep{wu2024word}) averaged across 6 translation directions on the WMT23 benchmark.}
    \label{fig:faithfulness_quality_trend}
\end{figure}

The field of Machine Translation (MT) has been revitalized by large language models (LLMs)~\citep{achiam2023gpt,chen2025benchmarking,lyu2025docia}. While Supervised Fine-Tuning (SFT) has achieved remarkable success~\citep{jiang2024mixtral,dong2025two}, its reliance on Maximum Likelihood Estimation (MLE) often fails to capture the nuances of human preference~\citep{xu2024paradigm,lu2024llamax,jiang2025survey}. Although Reinforcement Learning from Human Feedback (RLHF)~\citep{ouyang2022training} addresses this by aligning models with human values, its dependency on complex online interactions and training instability has shifted attention toward offline alternatives~\citep{she2024mapo,yang2025language,tan2025remedy}. Consequently, direct preference-based learning methods, such as Direct Preference Optimization (DPO)~\citep{rafailov2023direct} and Contrastive Preference Optimization (CPO)~\citep{xu2024contrastive}, efficiently align models with ``chosen'' versus ``rejected'' translations.

\begin{figure*}[t]
    \centering
    \includegraphics[width=1\textwidth]{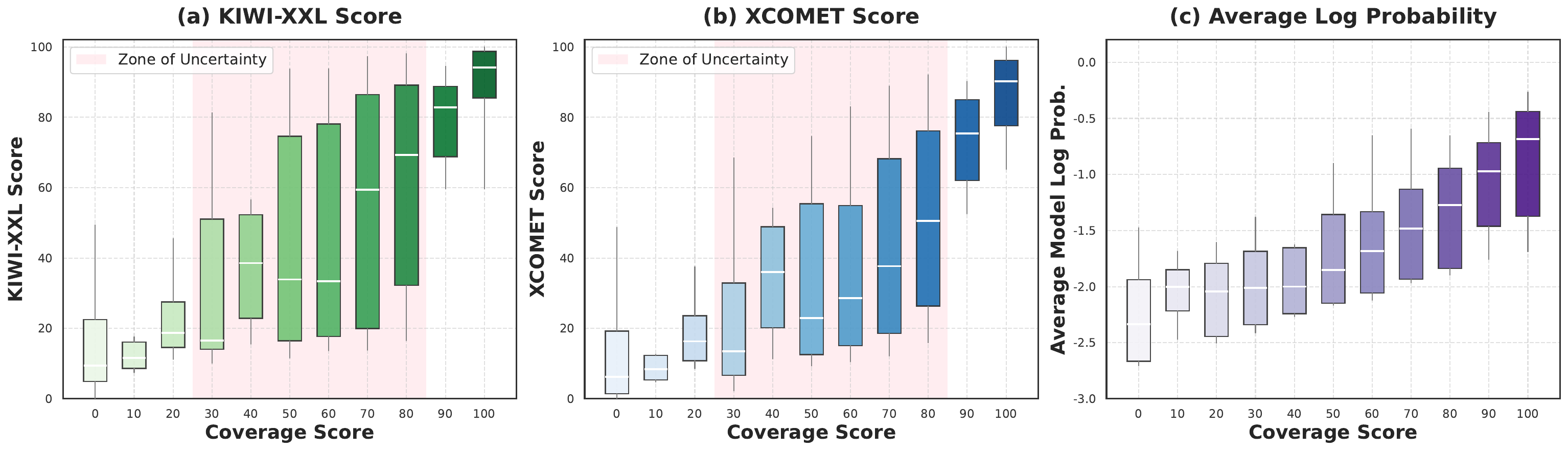}
    \caption{Analysis of \textbf{Metric Stability} across coverage levels (faithfulness) on the HalOmi benchmark. The plots illustrate the relationship between the Coverage Score and: (a) KIWI-XXL Score, (b) XCOMET Score, and (c) Sequence-level Average Model Log Probability (derived from the base model GemmaX2-28-9B). The shaded area highlights a ``Zone of Uncertainty'' (coverage range $30{-}80$), where partial errors are prevalent.}
    \label{fig:combined_metrics_vs_coverage}
\end{figure*}

Applying preference optimization to MT is constrained by reward reliability.
Neural \textbf{Quality Estimation (QE)} models~\citep{rei2023scaling,guerreiro2024xcomet} have replaced n-gram metrics (e.g., BLEU) as the preferred reward proxy due to their superior alignment with human perception.
However, corroborating findings on the limitations of QE metrics~\citep{deutsch2022limitations, dale2023detecting}, our analysis (Section~\ref{sec:motivation}) reveals a critical bias: QE models often over-prioritize surface-level well-formedness. Consequently, they fail to strictly penalize \textbf{partial errors} (exemplified in Appendix~\ref{app:qualitative_analysis}), assigning deceptively high scores to candidates that are superficially fluent but semantically flawed.
Compounding this, we find these errors concentrate in the \textit{intermediate faithfulness range} (Figure~\ref{fig:combined_metrics_vs_coverage}).
Crucially, even under a faithful ranking, this specific distribution renders them invisible to standard alignment strategies relying on single best-versus-worst pairs, which typically discard such non-extreme samples~\citep{he2024improving,zeng2024teaching,sun2025enhancing}.

To address these challenges, we introduce \textbf{M\textsuperscript{2}PO} (\textbf{M}ulti-Perspective \textbf{M}ulti-Pair \textbf{P}reference \textbf{O}ptimization), a data-centric framework designed to refine preference learning. Our framework consists of two complementary components. First, we employ \textit{Multi-Perspective Preference Construction} to mitigate the limitations of QE metrics by integrating a discrete faithfulness coefficient. This mechanism strictly penalizes translation errors while preserving general quality, further enhanced by a dynamic curriculum that fuses expert scores with the model's evolving confidence. Second, we propose \textit{Multi-Pair Joint Optimization} to maximize data utility by extending the standard pairwise objective to a dynamic multi-pair formulation. Through hierarchical contrasts, this strategy captures subtle errors, stabilized by a global ranking loss to regularize the model's output distribution.

We empirically validate M\textsuperscript{2}PO through extensive experiments on the WMT23, WMT24, and FLORES-200 benchmarks across six translation directions, applying our framework to two distinct base models (Qwen3-4B-Instruct~\citep{yang2025qwen3} and GemmaX2-28-9B~\citep{cui2025multilingual}). 
As illustrated in Figure~\ref{fig:faithfulness_quality_trend}, M\textsuperscript{2}PO effectively propels open-source models to a new performance frontier, achieving high faithfulness and quality simultaneously. 
Most notably, despite its modest parameter scale, our \textbf{GemmaX2-28-9B-M\textsuperscript{2}PO} not only surpasses its data generator (GPT-4o-mini~\citep{menick2024gpt}) and significantly larger open-source baselines (e.g., 30B+), but also demonstrates capabilities comparable to---and in some cases outperforming---powerful proprietary systems like GPT-4o~\citep{hurst2024gpt} and Gemini-2.0-Flash~\citep{comanici2025gemini}. We summarize our main contributions as follows:
\begin{itemize}[leftmargin=*,topsep=0.1em,itemsep=0.1em,parsep=0.1em]
    \item \textbf{Blind Spots in QE Rewards:} We identify a critical flaw in current QE metrics: they often prioritize fluency over faithfulness, providing deceptive signals that lead to reward hacking.
    
    \item \textbf{The M\textsuperscript{2}PO Framework:} We propose M\textsuperscript{2}PO, fusing multi-perspective alignment with multi-pair optimization. Diverging from standard pairwise methods, it leverages the full quality spectrum for maximal data efficiency.
    
    \item \textbf{Closing the Proprietary Gap:} M\textsuperscript{2}PO enables a 9B model to rectify supervision noise, surpassing its data generator (GPT-4o-mini) and matching proprietary systems (GPT-4o, Gemini-2.0-Flash).
\end{itemize}

\section{Related Work}

\subsection{LLM-Based Translation and Alignment}

While SFT builds foundational capabilities~\citep{rei2025tower+,wu2025calibrating}, alignment is pivotal for fine-grained quality. RLHF methodologies~\citep{ouyang2022training} diverge into online tracks like PPO~\citep{schulman2017proximal}, GRPO~\citep{guo2025deepseek}, and GSPO~\citep{zheng2025group}, which maximize rewards via iterative interactions~\citep{ramos2024fine,feng2025mtZero,he2025r1,li2025rival}, and cost-effective offline implicit variants~\citep{ahmadian2024multilingual,yang2025implicit,feng2025mtTree}. Recent advances in implicit optimization primarily focus on two parallel dimensions: signal construction and data efficiency. For the former, studies integrate QE metrics to scale supervision signals~\citep{xu2024x,he2024improving,agrawal2024modeling}. 
In parallel, to address data efficiency, Set-MPO~\citep{gupta2024multi} and LiPO~\citep{liu2024lipo} exploit the candidate spectrum via set-level contrasts and listwise ranking, respectively. 
However, their direct application to MT relies on the assumption of reliable rewards. 
Consequently, they remain susceptible to the \textit{deceptive reward signals}, where fluent yet unfaithful translations mislead metrics.

\subsection{Hallucination and Omission Mitigation}
Ensuring faithfulness remains a central challenge in LLM-based MT, primarily manifesting through two distinct failure modes \citep{zhang2024paying}.
\textbf{Hallucinations} involve the fabrication of content not present in the source~\citep{guerreiro2023hallucinations,himmi2024enhanced,gogoulou2025can}, whereas \textbf{omissions} fail to convey critical source information despite the output's grammatical correctness~\citep{yang2019reducing,vamvas2022little,dale2023halomi}.
Both errors are deceptive, as they are often masked by the model's high linguistic fluency.
To address unfaithfulness, recent works like WAP~\citep{wu2024word} leverage word alignment signals as constraints; however, such approaches often compromise general translation quality (e.g., lower COMET scores) in exchange for stricter alignment.

\begin{figure*}[t]
    \centering
    \includegraphics[width=1\textwidth]{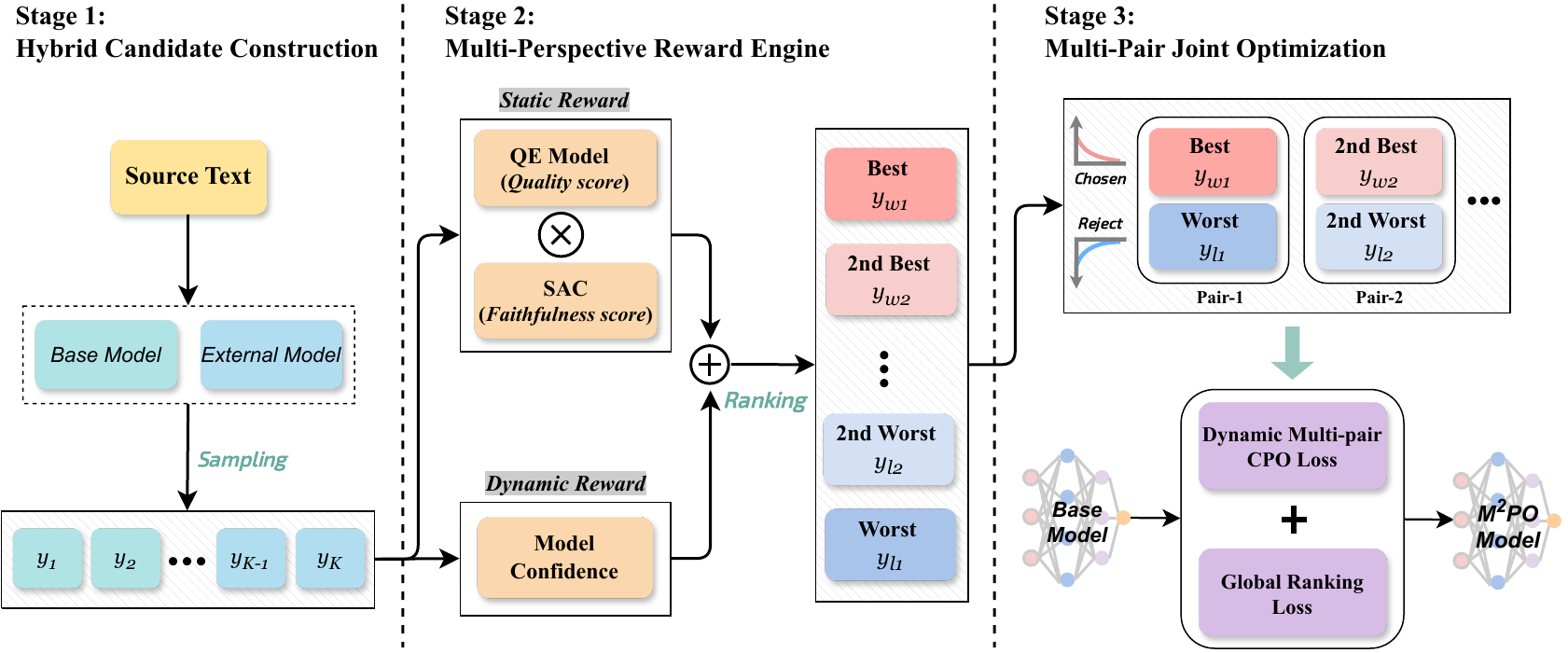}
    \caption{The M\textsuperscript{2}PO framework. The pipeline comprises: (1) Hybrid Candidate Construction for diverse sampling; (2) Multi-Perspective Reward Engine which ranks candidates by fusing static and dynamic signals; and (3) Multi-Pair Joint Optimization which updates the model using multiple contrastive pairs and a global ranking constraint.}
    \label{fig:framework}
\end{figure*}

\section{The ``Zone of Uncertainty'': QE Instability vs. Model Consistency}
\label{sec:motivation}

While QE models serve as scalable proxies for human judgment, they often over-prioritize surface-level fluency~\citep{deutsch2022limitations}, resulting in a systematic blind spot for nuanced errors—specifically, partial hallucinations and omissions, hereafter termed \textbf{partial errors}. To validate this, we analyze metric behavior on the HalOmi benchmark~\citep{dale2023halomi}. We contrast standard QE metrics, XCOMET-XXL\footnote{\url{https://huggingface.co/Unbabel/XCOMET-XXL}} and COMET-KIWI-XXL\footnote{\url{https://huggingface.co/Unbabel/wmt23-cometkiwi-da-xxl}} (hereafter \textbf{XCOMET} and \textbf{KIWI-XXL}), against \textbf{Coverage Score}---an LLM-based metric we implement using Gemini-2.0-Flash following the methodology of~\citet{wu2024word}, which quantifies the proportion of source semantic units preserved in the translation.
Consistent with prior findings of~\citet{wu2024word}, our analysis (Appendix~\ref{app:metric_reliability}) further validates Coverage Score as a reliable reference for faithfulness in this study.

\paragraph{The ``Zone of Uncertainty'' in QE Metrics.}
Figure~\ref{fig:combined_metrics_vs_coverage} illustrates a sharp contrast in metric stability. Standard QE metrics effectively secure the boundaries, showing high consistency in both the Low-Score Region (overt failures) and High-Score Region (near-perfect translations).
However, they falter in the critical interval ($30{-}80$)—denoted as the \textbf{``Zone of Uncertainty.''}
In this zone (dominated by partial errors), metrics lose their monotonic alignment with faithfulness (Panels a--b), frequently assigning misleadingly high scores to flawed candidates—echoing the critical blind spot for fluent hallucinations identified by \citet{dale2023detecting}.
Detailed analysis in Appendix~\ref{app:metric_reliability} corroborates that while QE excels at the extremes, it lacks the granularity to differentiate subtle failures within this unstable middle ground.

\paragraph{Model Confidence as a Stabilizer.}
Inspired by \citet{dale2023detecting}, who posit that model internal states outperform external metrics in detecting hallucinations, we validate this hypothesis on the HalOmi benchmark.
In contrast to the volatility of QE, Figure~\ref{fig:combined_metrics_vs_coverage}(c) reveals that the sequence-level average log probability of the base model (GemmaX2-28-9B) maintains a robust monotonic trajectory across the coverage spectrum.
Crucially, even when unfaithful translations appear linguistically fluent, the model's internal probability distribution reveals underlying uncertainty (``lying with hesitation'').
This empirical evidence confirms that the model's internal belief can serve as a reliable regularizer against external reward blind spots, a key insight driving our M\textsuperscript{2}PO framework.


\section{\texorpdfstring{The M\textsuperscript{2}PO Framework}{The M²PO Framework}} \label{sec:method}

We introduce \textbf{M\textsuperscript{2}PO}, a data-centric framework addressing the reliability blind spots of QE metrics and the inefficiency of standard alignment. As illustrated in Figure~\ref{fig:framework}, the framework is organized into two primary logical components structuring our discussion: \textit{Multi-Perspective Preference Construction} (covering Stages 1 \& 2) and \textit{Multi-Pair Joint Optimization} (Stage 3).

\subsection{Preliminaries}
\label{sec:preliminaries}
\paragraph{Preference Optimization Backbone.}
Let $\mathcal{D} = \{(x, y_w, y_l)\}$ denote a preference dataset, where $x$ represents the source input, and $y_w, y_l$ are the preferred and rejected candidate translations.
To mitigate the memory overhead of reference-model-dependent methods like DPO, we optimize the policy $\pi_\theta$ using the reference-model-free \textbf{CPO} paradigm, which enforces fidelity by synergizing preference alignment with generation stability:
\begin{equation}
\resizebox{1\hsize}{!}{$
\mathcal{L}_{\text{CPO}}(x, y_w, y_l) = \underbrace{- \log \sigma \Big( \beta \log \frac{\pi_{\theta}(y_w|x)}{\pi_{\theta}(y_l|x)} \Big)}_{\mathcal{L}_{\text{Pref}}} + \underbrace{ \big( - \log \pi_{\theta}(y_w|x) \big)}_{\mathcal{L}_{\text{NLL}}}
$}
\label{eq:cpo_unified}
\end{equation}
Here, $\mathcal{L}_{\text{Pref}}$ maximizes the likelihood margin between $y_w$ and $y_l$ scaled by $\beta$. Crucially, the negative log-likelihood term $\mathcal{L}_{\text{NLL}}$ acts as a regularizer, anchoring the model to high-quality responses to prevent the degradation of linguistic fluency.

\subsection{Multi-Perspective Preference Construction}
\label{subsec:prefer_constr}
We construct robust signals by integrating diverse sampling with multi-perspective scoring to capture the full spectrum of translation quality.

\paragraph{Hybrid Candidate Generation.}
To balance high-quality supervision with diverse error exposure, we construct a hybrid candidate pool $Y = \{y_1, \dots, y_K\}$ of even size $K$ for each source $x$.
We sample $K/2$ candidates from a strong external model and the remaining $K/2$ from the base model.
By uniformly applying high-randomness decoding ($T=0.8, \text{Top-}p=0.8$) to both sources, we aggressively diversify the candidate space.
The resulting quality spectrum—ranging from strict fidelity to fluent yet unfaithful partial errors—provides the critical contrast required for learning fine-grained boundaries based solely on semantic alignment.

\paragraph{Calibrating QE with Semantic Alignment Classifier.}
Reliance on standard QE metrics for reward scoring often yields misleading signals within the ``Zone of Uncertainty.'' To address this limitation, we prioritize LLM-based signals over traditional word alignment (e.g., WSPAlign~\citep{wu2023wspalign}), as our analysis (Appendix~\ref{app:metric_reliability}) highlights the latter's inefficiency in capturing nuanced partial errors. We introduce the \textit{Semantic Alignment Classifier} (\textbf{SAC}), an LLM-driven discriminator that leverages discrete diagnostics to robustly rectify QE biases. SAC maps candidates into four tiers: \textit{Full Match} ($1.0$), \textit{High Partial} ($0.7$), \textit{Low Partial} ($0.3$), and \textit{No Match} ($0.1$), where the lowest tier serves as a safety floor to prevent signal erasure. Implementation details and sensitivity analyses are provided in Appendix~\ref{app:sac_details}.
We formulate the static reward $r_s$ via multiplicative gating:
\begin{equation}
r_{s}(x, y_k) = r_{qe}(x, y_k) \cdot S_{align}(x, y_k)
\label{eq:static_score_multiplicative}
\end{equation}
where $r_{qe} \in [0, 100]$ is the KIWI-XXL score and $S_{align}$ is the corresponding SAC coefficient.
This formulation acts as a severity-aware veto, suppressing high-confidence yet unfaithful candidates (the ``Zone of Uncertainty'') towards the safety floor to ensure strict penalization and stability.

\paragraph{Dual-Perspective Fusion via Dynamic Curriculum.}
We construct a holistic reward by fusing the static faithfulness score $r_s(x, y_k)$ (Eq.~\ref{eq:static_score_multiplicative}) with the model's dynamic intrinsic confidence $r_d(x, y_k) = \frac{1}{|y_k|} \log \pi_{\theta}(y_k | x)$.
To harmonize the bounded $r_s$ and unbounded $r_d$, we apply z-score normalization over the candidate set $Y$ via $\hat{r} = (r - \mu_{Y})/\sigma_{Y}$.
These standardized terms are integrated via a dynamic weighting scheme:
\begin{equation}
r_{\text{fused}}(x, y_k) = \sigma \left( (1-\alpha_t)\hat{r}_s + \alpha_t \hat{r}_d \right)
\label{eq:fused_score}
\end{equation}
where $\sigma(\cdot)$ denotes the sigmoid function used for normalization. We adopt a linear curriculum by increasing $\alpha_t \in [0.1, 0.9]$ over the training steps $t$. This transitions the optimization from external grounding toward self-refinement, with the $0.9$ upper bound preserving sufficient external supervision to prevent reward over-optimization.

\begin{figure}[t]
    \centering
    \includegraphics[width=0.95\linewidth]{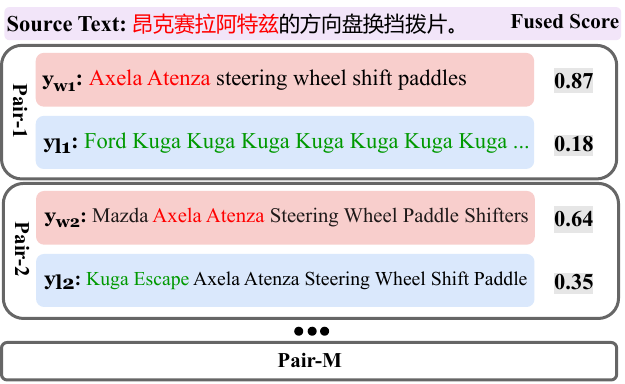}
\caption{\label{fig:case_visual} Visual comparison of Multi-Pair optimization (Zh$\rightarrow$En). \textbf{Pair-1} shows a wide margin driven by a severe error, while \textbf{Pair-2} presents a ``hard negative'' (partial error) with a narrower margin. The optimization extends to \textbf{Pair-M}, corresponding to the $K/2$ split.}
\end{figure}

\subsection{Multi-Pair Joint Optimization}
To maximize data efficiency, we propose a Dynamic Multi-Pair Strategy that transcends standard best-vs-worst comparisons. 
As illustrated in Figure~\ref{fig:case_visual}, while traditional methods rely on the sharpest distinction (Pair-1), they neglect the intermediate ``Zone of Uncertainty.'' 
M\textsuperscript{2}PO addresses this by mining the full candidate spectrum. 
We sort candidates by descending $r_{\text{fused}}$ to obtain the ranked set $\{y_{(1)}, \dots, y_{(K)}\}$ and construct $M=K/2$ pairs by coupling the top-$i$ and bottom-$i$ candidates: $(y_{w}^i, y_{l}^i) = (y_{(i)}, y_{(K-i+1)})$.
This hierarchical coupling creates a gradient of difficulty, ranging from clear quality gaps to subtle ``hard negatives'' (e.g., Pair-2) that demand deeper discrimination.

\paragraph{Local: Dynamic Multi-Pair CPO.}
Building upon these hierarchical contrasts, we minimize a joint preference-NLL objective, weighting each pair by its normalized margin gap $\Delta r^i = r_{\text{fused}}(x, y_{w}^i) - r_{\text{fused}}(x, y_{l}^i)$:
\begin{equation}
    \resizebox{0.88\hsize}{!}{$
    \begin{aligned}
        \mathcal{L}_{\text{DM-CPO}} &= \sum_{i=1}^{M} \left( \frac{\Delta r^i}{\sum_{j=1}^{M} \Delta r^j} \right) \cdot \mathcal{L}_{\text{CPO}}(x, y_{w}^i, y_{l}^i) 
    \end{aligned}
    $}
    \label{eq:dm_cpo_loss}
\end{equation}
This weighting anchors stability by emphasizing distinct quality gaps, while attenuating ambiguous pairs to prevent overfitting in the ``Zone of Uncertainty.''

\paragraph{Global: Ranking Regularization.}
To complement local pairwise discrimination, we enforce global ordinal consistency via a Listwise Ranking Loss~\citep{cao2007learning}:
\begin{equation}
    \mathcal{L}_{\text{Rank}} = - \sum_{k=1}^{K} P_{\text{fused}}(y_k|x) \log P_{\theta}(y_k|x)
    \label{eq:rank_loss}
\end{equation}
where the target distribution $P_{\text{fused}}(y_k|x)$ is defined as $\text{softmax}(r_{\text{fused}}(x, y_k)/\tau)$, and the model distribution $P_{\theta}(y_k|x)$ is derived by re-normalizing the policy log-probabilities $\log \pi_{\theta}(y_k|x)$ over the candidate set $Y$ via softmax.

\paragraph{Joint Objective.}
The final objective sums these components with weighting hyperparameter $\lambda_{\text{rank}}$:
\begin{equation}
\mathcal{L}_{\text{M\textsuperscript{2}PO}} = \mathcal{L}_{\text{DM-CPO}} + \lambda_{\text{rank}}\mathcal{L}_{\text{Rank}}
\end{equation}

\definecolor{c_closed}{HTML}{FCE4D6} 
\definecolor{c_open}{HTML}{FFF2CC}   
\definecolor{c_ours}{HTML}{E2EFDA}   

\begin{table*}[t]
\centering
\resizebox{\textwidth}{!}{%
\setlength{\tabcolsep}{3.5pt}
\renewcommand{\arraystretch}{1.07} 
\begin{tabular}{lccccccc}
\toprule
\textbf{Model} & \textbf{AVG} & \textbf{En$\to$Zh} & \textbf{En$\to$De} & \textbf{En$\to$Ja} & \textbf{Zh$\to$En} & \textbf{De$\to$En} & \textbf{Ja$\to$En} \\
\midrule
\cellcolor{c_closed}Gemini-2.0-Flash & 97.58 / 89.52 & 97.25 / 88.99 & 97.94 / \bb{89.13} & 97.65 / 91.09 & 97.67 / 93.15 & 97.95 / 86.66 & 97.01 / 88.12 \\
\cellcolor{c_closed}GPT-4o & \bb{97.89} / 89.16 & 97.36 / 87.96 & \bb{98.75} / 88.20 & 97.84 / 90.25 & 98.09 / 92.96 & 98.08 / 87.05 & 97.20 / 88.55 \\
\cellcolor{c_closed}GPT-4o-mini & 97.71 / 88.74 & 97.40 / 87.19 & 98.50 / 86.88 & 97.95 / 90.15 & 97.85 / 93.09 & 97.72 / 86.80 & 96.85 / 88.35 \\
\midrule

\cellcolor{c_open}Aya-expanse-32B & 97.06 / 88.70 & 96.96 / 88.50 & 97.71 / \cc{86.51} & 97.33 / 90.75 & 97.03 / 93.34 & 97.09 / 85.93 & 96.21 / 87.18 \\
\cellcolor{c_open}Aya-23-35B & 96.20 / 86.59 & 96.03 / 86.67 & 97.52 / 84.62 & 96.32 / 88.76 & 96.08 / 90.11 & 96.84 / 84.73 & 94.39 / 84.65 \\
\cellcolor{c_open}TowerInstruct-13B & 95.67 / 85.78 & 96.17 / 86.99 & 97.20 / 83.90 & 94.97 / 81.56 & 96.22 / 91.66 & 97.17 / 85.90 & 92.28 / 84.64 \\
\cellcolor{c_open}ALMA-13B-R & 95.13 / 85.30 & 95.63 / 86.63 & 96.85 / 83.97 & 93.99 / 79.84 & 95.79 / 92.12 & 96.92 / 85.55 & 91.60 / 83.70 \\
\midrule

\cellcolor{c_ours}Qwen3-4B-Instruct & 95.72 / 85.16 & 95.88 / 85.65 & 95.77 / 79.76 & 94.96 / 85.70 & 96.67 / 91.25 & 96.88 / 83.81 & 94.17 / 84.77 \\
\cellcolor{c_ours}\hspace{1em}+ SFT & 96.76 / 86.20 & 97.02 / 87.34 & 96.86 / 80.13 & 95.49 / 86.83 & 97.47 / 92.43 & 97.97 / 84.62 & 95.76 / 85.82 \\
\cellcolor{c_ours}\hspace{1em}+ CPO & 96.94 / 86.57 & 97.05 / 87.81 & 97.08 / 80.79 & 96.07 / 87.12 & 97.53 / 92.67 & 98.02 / 84.79 & 95.86 / 86.23 \\
\cellcolor{c_ours}\hspace{1em}\textbf{+ M\textsuperscript{2}PO} & 97.50 / 87.14 & 97.23 / 88.84 & 97.76 / 81.27 & 97.13 / 87.72 & \bbcc{98.25} / 92.89 & \bbcc{98.43} / 85.12 & 96.17 / 86.98 \\
\hline

\cellcolor{c_ours}GemmaX2-28-9B & 96.09 / 87.16 & 96.02 / 87.59 & 96.89 / 85.21 & 96.61 / 87.86 & 96.37 / 91.60 & 96.36 / 85.70 & 94.26 / 85.01 \\
\cellcolor{c_ours}\hspace{1em}+ SFT & 96.81 / 87.79 & 96.64 / 88.18 & 97.49 / 85.47 & 97.40 / 88.84 & 97.07 / 91.94 & 97.34 / 86.32 & 94.93 / 86.00 \\
\cellcolor{c_ours}\hspace{1em}+ Set-MPO\textsuperscript{\dag} & 96.90 / 88.06 & 96.48 / 87.99 & 97.28 / 85.67 & 97.75 / 89.12 & 97.19 / 92.26 & 96.98 / 86.20 & 95.73 / 87.11 \\
\cellcolor{c_ours}\hspace{1em}+ LiPO\textsuperscript{\dag} & 97.08 / 87.97 & 97.15 / 88.01 & 97.33 / 85.89 & 97.69 / 88.90 & 97.37 / 91.84 & 97.06 / 86.31 & 95.88 / 86.88 \\
\cellcolor{c_ours}\hspace{1em}+ CPO & 97.13 / 88.59 & 96.98 / 88.28 & 97.68 / 85.51 & 97.62 / 90.20 & 97.23 / 92.89 & 97.45 / 86.61 & 95.83 / 88.02 \\
\cellcolor{c_ours}\hspace{1em}\textbf{+ M\textsuperscript{2}PO} & \cc{97.87} / \bbcc{89.67} & \bbcc{97.55} / \bbcc{89.65} & \cc{98.13} / 86.21 & \bbcc{97.96} / \bbcc{92.16} & 97.98 / \bbcc{93.46} & 98.41 / \bbcc{87.56} & \bbcc{97.21} / \bbcc{88.98} \\
\bottomrule
\end{tabular}%
}

\caption{\setlength{\fboxsep}{1pt}
Main results on the WMT23 benchmark. Results are reported in the format of \textbf{Coverage / XCOMET}. Model names are color-coded by type: \colorbox{c_closed}{Proprietary}, \colorbox{c_open}{Open-source baselines}, and \colorbox{c_ours}{Our experiments}. \textbf{Bold} indicates the best result overall. \colorbox{mycolor}{Colored background} indicates the best result among open-source models. \textsuperscript{\dag} indicates methods originally proposed for general tasks, which we adapted and reproduced for machine translation.}
\label{tab:main_results}
\end{table*}

\section{Experiments}

\subsection{Datasets and Baselines}
\paragraph{Training Data: M\textsuperscript{2}PO-Prefer.}
We construct our training set, M\textsuperscript{2}PO-Prefer, using seeds derived from WMT22~\citep{kocmi2022findings}, IWSLT~\citep{cettolo2017overview}, and FLORES-200-dev~\citep{costa2022no}.
Following \S\ref{subsec:prefer_constr}, we construct the pool ($K=8$) with 4 candidates each from GPT-4o-mini (external model) and GemmaX2-28-9B (base model).
After rigorous filtering, the dataset comprises $\sim$20,000 source inputs (totaling $\sim$160,000 candidates) across six directions (En$\leftrightarrow$\{Zh, De, Ja\}), demonstrating remarkable cost-efficiency (see analysis in Appendix~\ref{subsec:cost_analysis}). Detailed dataset statistics are in Appendix~\ref{app:dataset_statistics}.

\paragraph{Test Benchmarks.}
We evaluate on official benchmarks strictly disjoint from training seeds: WMT23~\citep{kocmi2023findings} and FLORES-200-test~\citep{costa2022no} for bidirectional evaluation (En$\leftrightarrow$X), and WMT24~\citep{kocmi2024findings} for unidirectional En$\rightarrow$X assessment.

\paragraph{Comparison Systems.}
We benchmark M\textsuperscript{2}PO against three groups:
(1) \textbf{Proprietary Systems}: Gemini-2.0-Flash, GPT-4o, and GPT-4o-mini;\footnote{Versions: \texttt{gemini-2.0-\allowbreak flash-001}, \texttt{gpt-4o-\allowbreak 2024-\allowbreak 08-\allowbreak 06}, and \texttt{gpt-4o-\allowbreak mini-\allowbreak 2024-\allowbreak 07-\allowbreak 18}.}
(2) \textbf{Open-source LLMs}: Including general-purpose models (Aya-expanse-32B, Aya-23-35B)~\citep{dang2024aya, aryabumi2024aya} and translation-specialized models (TowerInstruct-13B, ALMA-13B-R)~\citep{alves2024tower, xu2024contrastive};
and (3) \textbf{Controlled Baselines}: Implementation of SFT (trained on the top-ranked candidates from M\textsuperscript{2}PO-Prefer) and preference-based methods (CPO, Set-MPO, LiPO) utilizing standard KIWI-XXL as the reward signal. Supplementary ablations on SFT gold references and CPO reward configurations are provided in Appendix~\ref{app:baseline_ablations} and \S\ref{sub:ablation_studies}, respectively.

\subsection{Implementation and Evaluation}
\paragraph{Model Configuration.}
We employ two advanced open-source base models: \textbf{GemmaX2-28-9B\footnote{\url{https://huggingface.co/ModelSpace/GemmaX2-28-9B-v0.1}}} and \textbf{Qwen3-4B-Instruct}\footnote{\url{https://huggingface.co/Qwen/Qwen3-4B-Instruct-2507}}. 
Both are adapted via LoRA~\citep{hu2022lora} ($r=32, \alpha=64$) and optimized using AdamW~\citep{loshchilov2017decoupled} with a learning rate of $5 \times 10^{-5}$, batch size of 32, and sequence length of 512, utilizing a 0.1 warmup ratio and 1.0 gradient clipping.
For M\textsuperscript{2}PO, we set $\beta=0.1$, $\lambda_{\text{rank}}=0.5$, and $\tau=1.0$. Our experiments run on an NVIDIA H100 GPU for 2 epochs, requiring approximately 2 hours.

\paragraph{Evaluation Protocol.}
Inference is conducted via vLLM~\citep{kwon2023efficient} ($T=0.3, \text{Top-}p=0.3$). We adopt a comprehensive dual-aspect metric suite:

\textbf{(1) Faithfulness:} We employ the \textbf{Coverage Score} (via Gemini-2.0-Flash~\citep{wu2024word}) as our primary faithfulness metric, supported by the discussion in \S\ref{sec:motivation}.

\textbf{(2) Translation Quality:} We employ the widely recognized \textbf{XCOMET} for reference-free evaluation. While we acknowledge QE blind spots for partial errors, our comprehensive analysis (\S\ref{sec:motivation} and Appendix~\ref{app:metric_reliability}) confirms that XCOMET remains a highly reliable discriminator \textit{within the high-faithfulness regime}. Thus, once faithfulness is secured, it effectively measures fluency and nuance. To mitigate potential bias, we complement this with reference-based \textbf{COMET-22}\footnote{\url{https://huggingface.co/Unbabel/wmt22-comet-da}}~\citep{rei2020comet}.

\begin{figure}[t]
    \centering
    \includegraphics[width=1\linewidth]{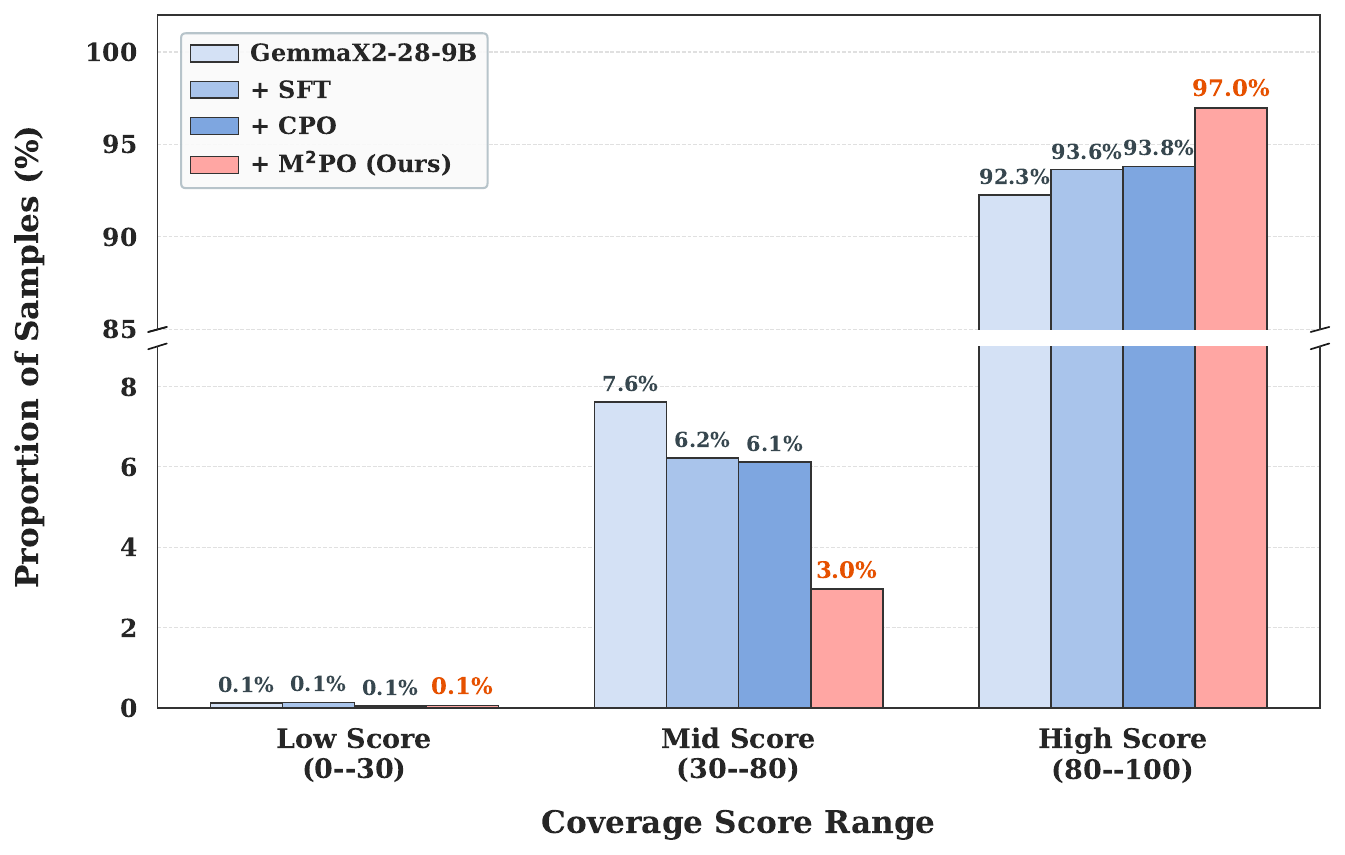}
    \caption{Distribution of sample proportions across Low (0--30), Mid (30--80), and High (80--100) Coverage Score intervals on the WMT23 benchmark.}
    \label{fig:dist_comparison}
\end{figure}

\subsection{Main Results}
Table~\ref{tab:main_results} presents the primary performance on WMT23, where M\textsuperscript{2}PO achieves leading results among open baselines, effectively closing the gap with proprietary systems.
Specifically, our \textit{GemmaX2-28-9B-M\textsuperscript{2}PO} outperforms all open baselines—including the significantly larger Aya-expanse-32B—by achieving a 97.87 Coverage Score and 89.67 XCOMET.
Notably, it surpasses its data generator (GPT-4o-mini) and even exceeds GPT-4o in translation quality (+0.51 XCOMET) while maintaining comparable faithfulness.
At smaller scales, \textit{Qwen3-4B-Instruct-M\textsuperscript{2}PO} demonstrates exceptional efficiency, surpassing 13B baselines and securing the highest Coverage on Zh$\to$En (98.25) and De$\to$En (98.43) among all systems.

Controlled comparisons further confirm the algorithmic superiority of our framework. M\textsuperscript{2}PO yields substantial gains over standard CPO (+1.08 XCOMET) as well as advanced set-level (Set-MPO) and listwise (LiPO) adaptations. This validates that the synergy of multi-perspective preference labeling and multi-pair optimization provides cleaner, more granular learning signals than general-purpose alignment methods.

Beyond standard benchmarks, we rigorously validate robustness (Appendix~\ref{app:additional_experiments}). 
First, consistent effectiveness on COMET-22 confirms that gains are not artifacts of specific metric optimization. 
Second, FLORES-200 and WMT24 evaluations verify generalization across diverse domains and temporal shifts, ruling out overfitting. 
Finally, qualitative case studies (Appendix~\ref{app:qualitative_analysis}) provide concrete evidence of M\textsuperscript{2}PO rectifying subtle hallucinations and omissions that persist in baselines.

\section{Analysis}

\subsection{Mitigating Partial Errors in the ``Zone of Uncertainty''}
\label{sec:distr_sample}
Figure~\ref{fig:dist_comparison} visualizes the full score distribution, situating the intermediate ``Zone of Uncertainty'' (30--80) between the low and high extremes.
A defining challenge in this regime is \textbf{sparsity}: deceptive partial errors comprise only ($\sim$6--7\%) of the data.
Instead of actively correcting them, standard optimization objectives tend to prioritize the majority of easier patterns, effectively diluting the learning signal from these rare, fine-grained failures.
Consequently, standard methods fail to mine these subtle signals; SFT and CPO plateau at comparable error rates (6.2\% and 6.1\%), offering limited gains over the base model.
In contrast, M\textsuperscript{2}PO explicitly amplifies these sparse instances via faithfulness-weighted penalties.
This acts as a precision filter, reducing partial errors by $\sim$51\% (to 3.0\%) and propelling the distribution into the high-faithfulness regime ($>$80 coverage), peaking at 97.0\%.
This confirms M\textsuperscript{2}PO rectifies fine-grained failures evading conventional alignment.

\begin{table}[t]
\centering
\renewcommand{\arraystretch}{1.1}
\resizebox{\linewidth}{!}{
\begin{tabular}{l|ccc|cc}
\toprule
\multirow{2}{*}{\textbf{System}} & \multicolumn{3}{c|}{\textbf{Quality Score}} & \multicolumn{2}{c}{\textbf{Average Metric}} \\
 & \textbf{Zh} & \textbf{De} & \textbf{Ja} & \textbf{Quality} & \textbf{Faith. (\%)} \\
\midrule
GPT-4o-mini & 4.11 & \textbf{4.32} & 4.56 & 4.33 & 87.11 \\
\midrule
GemmaX2-28-9B & 3.73 & 3.83 & 3.04 & 3.53 & 72.67 \\
\textbf{M\textsuperscript{2}PO (Ours)} & \textbf{4.14} & 4.30 & \textbf{4.67} & \textbf{4.37} & \textbf{89.11} \\
\bottomrule
\end{tabular}
}
\caption{Human evaluation results on 450 sampled WMT23 instances (150 per direction across En-Zh, En-De, En-Ja). \textbf{Quality} measures holistic fluency and adequacy (0--5); \textbf{Faith.} denotes the percentage of ``Full Match'' labels averaged across languages.}
\label{tab:human_eval}
\end{table}

\subsection{Human Evaluation Validation}
We conducted a blind expert evaluation on 450 representative WMT23 instances. To ensure balanced coverage of challenging cases, we employed stratified sampling across the ``Zone of Uncertainty'' and remaining regions. Annotators adhered to a dual-aspect protocol: (1) holistic Quality Ratings (0--5) adapted from WMT22 standards~\citep{kocmi2022findings}; and (2) manual Faithfulness Classification via our fine-grained SAC taxonomy.

As shown in Table~\ref{tab:human_eval}, M\textsuperscript{2}PO demonstrates superior performance across both dimensions. It achieves peak faithfulness (89.11\%), marking a substantial +16.4\% gain over the base model. Notably, it surpasses its data generator, GPT-4o-mini (87.11\%), acting as a \textit{denoising filter} to rectify subtle errors. In addition to exceptional faithfulness, M\textsuperscript{2}PO attains the highest holistic Quality Score (4.37), outperforming the proprietary baseline (4.33) and excelling in linguistically distant languages (e.g., 4.67 in En-Ja), thereby effectively reconciling the trade-off between faithfulness and fluency.

\begin{table}[t]
    \centering
    \small
    \renewcommand{\arraystretch}{1.1}
    \resizebox{\columnwidth}{!}{%
        \setlength{\tabcolsep}{10pt}
        \begin{tabular}{lcc}
        \toprule
        \textbf{Base Objective} & \textbf{Standard} & \textbf{+ M\textsuperscript{2}PO (Ours)} \\
        \midrule
        \textit{Base Model} & 96.09 / 87.16 & --- \\
        \midrule
        DPO &  96.91 / 88.48 & 97.54 / 89.15 \\
        KTO &  96.71 / 88.39 & 97.37 / 89.43 \\
        SimPO & 97.06 / \textbf{88.74} & 97.73 / 89.55 \\
        ORPO & 96.47 / 87.86 & 96.85 / 88.41 \\
        CPO & \textbf{97.13} / 88.59 & \textbf{97.87} / \textbf{89.67} \\
        \bottomrule
        \end{tabular}%
    }
    \caption{Generalization across different preference objectives on WMT23. ``+ M\textsuperscript{2}PO'' indicates plugging the standard loss (e.g., $\mathcal{L}_{\text{DPO}}$) into our framework by replacing $\mathcal{L}_{\text{CPO}}$ in Eq.~\ref{eq:dm_cpo_loss}. Scores are \textbf{Coverage / XCOMET}.}
    \label{tab:ablation_algorithms}
\end{table}

\subsection{Generalization Across Preference Objectives}
M\textsuperscript{2}PO is a flexible framework designed to enhance offline preference optimization. We validate its generality by extending it to diverse DPO-style objectives: DPO~\citep{rafailov2023direct}, KTO~\citep{ethayarajh2024kto}, SimPO~\citep{meng2024simpo}, and ORPO~\citep{hong2024orpo}.

\paragraph{Integration Mechanism.}
The ``+ M\textsuperscript{2}PO'' configuration represents a dual enhancement. First, we use our multi-perspective rewards to construct high-quality preference pairs. Second, we substitute the specific term $\mathcal{L}_{\text{CPO}}$ in Eq.~\ref{eq:dm_cpo_loss} with the target objective (e.g., $\mathcal{L}_{\text{DPO}}$) while retaining the dynamic weights. This allows standard baselines to benefit from both our refined data selection strategy and our dynamic calibration mechanism.

\paragraph{Performance.}
Table~\ref{tab:ablation_algorithms} shows consistent gains across all baselines. Enhancing the strongest baseline (CPO) yields peak performance (97.87 Coverage), while KTO sees a substantial boost of +1.04 XCOMET. Furthermore, we demonstrate that M\textsuperscript{2}PO remains competitive even against computation-heavy Online-RL algorithms (see Appendix~\ref{app:online_rl} for detailed comparisons).

\subsection{Generalization to Alternative Reward Proxies}
\label{sec:metric_robustness}

To evaluate the generalization of M\textsuperscript{2}PO, we substitute \textbf{KIWI-XXL} with \textbf{MetricX-24-XXL}\footnote{\url{https://huggingface.co/google/metricx-24-hybrid-xxl-v2p6}}, a leading regression-based evaluation metric. As shown in Table~\ref{tab:metric_generalization}, standalone MetricX-24-XXL (97.35) already outperforms standalone KIWI-XXL (96.99). However, integrating our SAC penalty with MetricX-24-XXL yields further gains (+0.37 Coverage). While our default KIWI-XXL + SAC configuration marginally achieves the highest overall score (97.87), the broader takeaway is that integrating SAC consistently improves the evaluated continuous metrics, resulting in comparable alignment efficacy across the hybrid configurations.

To understand why explicit faithfulness supervision remains vital even with stronger continuous metrics, we conduct a qualitative analysis of partial errors (detailed in Appendix~\ref{app:metric_comparison}). We observe that both MetricX-24-XXL and KIWI-XXL share a critical blind spot within the ``Zone of Uncertainty'' (introduced in \S\ref{sec:motivation}): \textbf{Ranking Inconsistency}. They tend to prioritize surface-level fluency over strict semantic faithfulness, occasionally assigning better scores to highly fluent outputs with major omissions than to those with minor errors. SAC mitigates this vulnerability by providing a robust, discrete penalty that explicitly corrects these ranking inversions. This confirms that continuous QE proxies (assessing overall translation quality) and SAC (acting as a strict semantic guardrail) are fundamentally complementary.

\begin{table}[t]
    \centering
    \footnotesize
    \renewcommand{\arraystretch}{1.2} 
    \setlength{\tabcolsep}{0pt}
    \begin{tabular*}{\linewidth}{@{\extracolsep{\fill}} l c }
    \toprule
    \textbf{Reward Configuration for M\textsuperscript{2}PO} & \textbf{Coverage} \\
    \midrule
    \textit{Base (GemmaX2-28-9B)} & 96.09 \\
    \midrule
    MetricX-24-XXL only & 97.35 \\
    MetricX-24-XXL + SAC & 97.72 \\
    KIWI-XXL only & 96.99 \\
    \textbf{KIWI-XXL + SAC (Ours)} & \textbf{97.87} \\
    \bottomrule
    \end{tabular*}
    \caption{Performance comparison on WMT23 across different reward proxies. We evaluate the impact of the SAC penalty when integrated with KIWI-XXL and MetricX-24-XXL.}
    \label{tab:metric_generalization}
\end{table}

\subsection{Ablation Studies}
\label{sub:ablation_studies}
We conduct systematic ablations on WMT23 (Table~\ref{tab:ablation}) to dissect M\textsuperscript{2}PO. To rigorously attribute performance gains, we structure the analysis into two dimensions: validating the fundamental optimization objectives and isolating the efficacy of specific algorithmic mechanisms.

\paragraph{Impact of Optimization Objectives.}
The DM-CPO engine ($\mathcal{L}_{\text{DM-CPO}}$) acts as the cornerstone; its removal results in the sharpest decline (-1.11 Coverage and -1.58 XCOMET), confirming its primacy in enforcing semantic adherence. Complementarily, excluding the Ranking Loss ($\mathcal{L}_{\text{Rank}}$) compromises global ordinal consistency, validating that global anchoring serves as a necessary safeguard against faithfulness degradation (see Appendix~\ref{app:lambda_rank_analysis} for weight sensitivity analysis).

\paragraph{Impact of Core Strategies.}
We further validate specific algorithmic mechanisms.
(1) SAC Fidelity Filter: Removing this constraint ($S_{align} = 1$) drops Coverage to 96.99, confirming it suppresses fluent errors typically overlooked by standard QE (sensitivity analysis in Appendix~\ref{app:sac_details}).
(2) Dynamic Fusion: Ablating the curriculum ($\alpha_t = 0$) impairs performance, underscoring the need to balance \textit{external grounding} with \textit{intrinsic confidence} to mitigate static proxy biases (sensitivity analysis in Appendix~\ref{app:static_vs_dynamic}).
(3) Multi-Pair Contrast: Removing this component reduces our method to a standard single-pair CPO baseline trained with our augmented reward (KIWI-XXL + SAC + Model Confidence). The consistent performance drop across all metrics proves that contrasting multiple candidates provides a clearer learning signal than relying on a single positive-negative pair.

\begin{table}[t]
    \centering
    \footnotesize 
    \renewcommand{\arraystretch}{1.05} 
    \setlength{\tabcolsep}{0pt}       
    \begin{tabular*}{\linewidth}{@{\extracolsep{\fill}} l cc } 
    \toprule
    \textbf{Model Config.} & \textbf{Coverage} & \textbf{XCOMET} \\
    \midrule
    \textbf{Full M\textsuperscript{2}PO} & \textbf{97.87} & \textbf{89.67} \\ 
    \midrule
    \multicolumn{3}{l}{\textit{Impact of Loss Components}} \\
    \hspace{1em} w/o $\mathcal{L}_{\text{DM-CPO}}$ & 96.76 & 88.09 \\
    \hspace{1em} w/o $\mathcal{L}_{\text{Rank}}$ & 97.15 & 89.19 \\
    \midrule
    \multicolumn{3}{l}{\textit{Impact of Core Strategies}} \\
    \hspace{1em} w/o SAC Fidelity Filter & 96.99 & 89.28 \\ 
    \hspace{1em} w/o Dynamic Fusion & 97.06 & 89.24 \\
    \hspace{1em} w/o Multi-Pair & 97.42 & 89.16 \\ 
    \midrule
    \textit{Base (GemmaX2-28-9B)} & 96.09 & 87.16 \\ 
    \bottomrule
    \end{tabular*}
    
    \caption{Ablation studies on the WMT23 benchmark.}
    \label{tab:ablation}
\end{table}

\subsection{Independence from Proprietary Distillation}
\label{sec:distillation_ablation}

A critical question is whether M\textsuperscript{2}PO's gains stem primarily from proprietary teacher distillation (e.g., GPT-4o-mini) or the proposed multi-pair optimization framework itself. To disentangle this, we compare an \textit{Open-Source Pipeline} against a \textit{Proprietary Pipeline}.

The \textit{Open-Source Pipeline} restricts candidate generation to the base model (GemmaX2-28-9B). Because this base model is specialized for translation and lacks robust evaluation capabilities, we utilize Gemma-3-27B-it\footnote{\url{https://huggingface.co/google/gemma-3-27b-it}} for SAC labeling. Conversely, the \textit{Proprietary Pipeline} relies entirely on GPT-4o-mini for both candidate generation and labeling.

As shown in Table~\ref{tab:distillation_ablation}, the open-source M\textsuperscript{2}PO consistently outperforms the CPO baseline (+0.63 Coverage, +0.74 XCOMET). Furthermore, removing the SAC signal degrades performance (to 96.83 Coverage and 88.73 XCOMET), mirroring the ablation trends of the proprietary setting. These results confirm that M\textsuperscript{2}PO's improvements are intrinsically driven by its multi-pair optimization and multi-reward signals, rather than simply acting as a distillation vehicle. Crucially, this validates a fully open-source paradigm for deploying M\textsuperscript{2}PO without reliance on commercial APIs.

\begin{table}[t]
    \centering
    \footnotesize 
    \renewcommand{\arraystretch}{1.05} 
    \setlength{\tabcolsep}{0pt} 
    \begin{tabular*}{\linewidth}{@{\extracolsep{\fill}} l cc }
    \toprule
    \textbf{Model Configuration} & \textbf{Coverage} & \textbf{XCOMET} \\
    \midrule
    \textit{Base (GemmaX2-28-9B)} & 96.09 & 87.16 \\
    \midrule
    \multicolumn{3}{l}{\textit{Open-Source Pipeline}} \\
    \hspace{1em} + CPO & 96.74 & 88.42 \\
    \hspace{1em} + M\textsuperscript{2}PO (w/o SAC) & 96.83 & 88.73 \\
    \hspace{1em} \textbf{+ M\textsuperscript{2}PO (w/ Open-Source SAC)} & \textbf{97.37} & \textbf{89.16} \\
    \midrule
    \multicolumn{3}{l}{\textit{Proprietary Pipeline}} \\
    \hspace{1em} + CPO & 97.13 & 88.59 \\
    \hspace{1em} + M\textsuperscript{2}PO (w/o SAC) & 96.99 & 89.28 \\
    \hspace{1em} \textbf{+ M\textsuperscript{2}PO (w/ Proprietary SAC)} & \textbf{97.87} & \textbf{89.67} \\
    \bottomrule
    \end{tabular*}
    \caption{Ablation study on WMT23 evaluating the fully open-source framework. The \textit{Open-Source Pipeline} uses the base model for candidate generation and Gemma-3-27B-it for SAC labeling, while the \textit{Proprietary Pipeline} uses GPT-4o-mini for both tasks.}
    \label{tab:distillation_ablation}
\end{table}

\section{Conclusion}

We present M\textsuperscript{2}PO, a data-centric framework addressing the blind spots of standard reward models and the data inefficiency in MT preference optimization. By integrating a multi-perspective alignment curriculum with a multi-pair strategy augmented by listwise ranking, M\textsuperscript{2}PO effectively targets the ``Zone of Uncertainty,'' rectifying deceptive partial errors overlooked by conventional methods. Extensive evaluations across WMT23, WMT24, and FLORES-200 confirm the framework's efficacy on 4B and 9B models, showing consistent gains in both faithfulness and translation quality. Results highlight that precise preference alignment enables compact open-source models to rival larger proprietary systems, establishing M\textsuperscript{2}PO as a robust pathway toward high-fidelity MT.

\section{Limitations}

While M\textsuperscript{2}PO demonstrates strong performance on compact models (4B/9B) across major translation directions, validating its scalability to larger architectures (e.g., 70B+) and low-resource languages remains an important direction for future work. Regarding the data and evaluation pipeline, our primary experiments rely on proprietary models (GPT-4o-mini, Gemini-2.0-Flash) to provide high-fidelity training signals and robust assessment. Although this reliance on commercial APIs may limit the immediate reproducibility of the full workflow, we have taken initial steps toward mitigation by validating a fully open-source alternative (\S\ref{sec:distillation_ablation}). Future work will further optimize this offline-deployable framework to narrow the gap with proprietary distillation and extend it to broader translation scenarios.

\section*{Acknowledgments}

The authors would like to thank the anonymous reviewers for their valuable and constructive feedback. This work was supported by Alibaba Group.

\bibliography{custom}


\appendix

\begin{table*}[t]
\centering
\small
\renewcommand{\arraystretch}{1.4}
\setlength{\tabcolsep}{4pt}

\begin{tabularx}{\textwidth}{m{2.5cm} >{\raggedright\arraybackslash}X c c c c}
\toprule
\multirow{2}{*}{\textbf{System}} & \multirow{2}{*}{\textbf{Translation Output}} & \multicolumn{2}{c}{\textbf{Training}} & \multicolumn{2}{c}{\textbf{Testing}} \\
\cmidrule(lr){3-4} \cmidrule(lr){5-6}
 & & \scriptsize \textbf{KIWI-XXL} & \scriptsize \textbf{SAC} & \scriptsize \textbf{XCOMET} & \scriptsize \textbf{Coverage} \\
\midrule
\multicolumn{6}{l}{\cellcolor{gray!15}\textbf{Source (zh$\to$en):} \cn{\textcolor{blue}{口水鸡}应该是熟的，但收到的是生肉，没办法吃}} \\
GemmaX2-28-9B & \textcolor{highpartial}{The cold chicken} should be cooked, but what I received was raw meat, making it inedible. 
& 62.01 & \textcolor{highpartial}{\textbf{High Partial}} & 85.57 & 85 \\
GPT-4o-mini & \textcolor{fullmatch}{The mouth-watering chicken} should be cooked, but what I received was raw meat, which was inedible. 
& 54.49 & \textcolor{fullmatch}{\textbf{Full match}} & 84.04 & 100 \\
M\textsuperscript{2}PO & \textcolor{fullmatch}{Poached chicken with chili sauce} should be cooked, but what I received was raw meat, making it inedible. 
& 63.48 & \textcolor{fullmatch}{\textbf{Full match}} & 89.52 & 100 \\
\midrule

\multicolumn{6}{l}{\cellcolor{gray!15}\textbf{Source (en$\to$zh):} At the time, nearly 100 residents were \textcolor{blue}{evacuated from the area}.} \\
GemmaX2-28-9B & \cn{当时，近100名居民从该地区被疏} \textcolor{lowpartial}{evacuated}\cn{。}
& 80.08 & \textcolor{lowpartial}{\textbf{Low Partial}} & 99.00 & 85 \\
GPT-4o-mini & \cn{当时，几乎有100名居民\textcolor{fullmatch}{从该地区撤离}。}
& 73.88 & \textcolor{fullmatch}{\textbf{Full match}} & 99.20 & 100 \\
M\textsuperscript{2}PO & \cn{当时，近100名居民被\textcolor{fullmatch}{从该地区疏散}。}
& 96.29 & \textcolor{fullmatch}{\textbf{Full match}} & 99.98 & 100 \\
\midrule

\multicolumn{6}{l}{\cellcolor{gray!15}\textbf{Source (de$\to$en):} Entsorgung: Bitte \textcolor{blue}{nur restentleerte} Gebinde dem Recycling zuführen.} \\
GemmaX2-28-9B & Disposal: Please submit \textcolor{highpartial}{emptied} containers for recycling. 
& 69.38 & \textcolor{highpartial}{\textbf{High Partial}} & 85.52 & 75 \\
GPT-4o-mini & Disposal: Please \textcolor{highpartial}{only} contribute \textcolor{highpartial}{emptied} containers to recycling. 
& 45.48 & \textcolor{fullmatch}{\textbf{Full match}} & 87.96 & 95 \\
M\textsuperscript{2}PO & Disposal: Please \textcolor{fullmatch}{only} submit \textcolor{fullmatch}{completely emptied} containers to recycling. 
& 76.71 & \textcolor{fullmatch}{\textbf{Full match}} & 89.52 & 100 \\

\bottomrule
\end{tabularx}
\caption{Case study: Comparison of translation alignment and quality metrics. \textbf{KIWI-XXL} and \textbf{SAC} serve as training signals, while \textbf{XCOMET} and \textbf{Coverage} are testing metrics. \textcolor{blue}{Blue} text marks the source segments of interest. In the Translation Output, \textcolor{fullmatch}{green} denotes faithful translations, whereas \textcolor{highpartial}{orange} and \textcolor{lowpartial}{red} visualize partial errors that standard QE metrics often overlook.}
\label{tab:case_study}
\end{table*}

\begin{table*}[t]
    \centering
    \small
    \resizebox{\textwidth}{!}{%
        \renewcommand{\arraystretch}{1.2}
        
        \begin{tabular}{l ccc ccc c ccc ccc}
        \toprule
        \multirow{3}{*}{\textbf{Metric}} & \multicolumn{6}{c}{\textbf{Score Distribution (Avg. Score $\uparrow$)}} & & \multicolumn{6}{c}{\textbf{Correlation Robustness (Pearson $\times 100$ $\uparrow$)}} \\
        \cmidrule(lr){2-7} \cmidrule(lr){9-14}
        & \multicolumn{3}{c}{Hallucination} & \multicolumn{3}{c}{Omission} & & \multicolumn{3}{c}{Hallucination} & \multicolumn{3}{c}{Omission} \\
        \cmidrule(lr){2-4} \cmidrule(lr){5-7} \cmidrule(lr){9-11} \cmidrule(lr){12-14}
        & No & Part. & Full & No & Part. & Full & & All & w/o Part. & $\Delta$ & All & w/o Part. & $\Delta$ \\
        \midrule
        
        \rowcolor{gray!15} 
        \multicolumn{14}{l}{\textit{\textbf{Word Alignment Baselines}}} \\
        WSPAlign & 64.4 & 45.9 & 25.2 & 69.3 & 51.2 & 25.6 & & 61.97 & 64.86 & +2.89 & 63.70 & 72.63 & +8.94 \\
        \midrule
        \rowcolor{gray!15} 
        \multicolumn{14}{l}{\textit{\textbf{Training Signals}}} \\
        SAC & 54.3 & 88.1 & 76.7 & 64.6 & 77.9 & 77.8 & & 73.99 & 75.59 & +1.60 & 71.22 & 78.54 & +7.32 \\
        KIWI-XXL & 63.7 & 24.3 & 9.8 & 64.3 & 36.1 & 9.4 & & 61.37 & 64.85 & \textbf{+3.48} & 54.27 & 65.87 & \textbf{+11.60} \\
        \midrule

        \rowcolor{gray!15} 
        \multicolumn{14}{l}{\textit{\textbf{Testing Metrics}}} \\
        Coverage & 85.0 & 48.5 & 9.3 & 81.0 & 60.5 & 8.7 & & \textbf{76.65} & \textbf{77.12} & +0.47 & \textbf{71.74} & \textbf{79.69} & +7.95 \\
        XCOMET   & 76.8 & 36.4 & 16.4 & 79.5 & 52.9 & 16.1 & & 69.72 & 72.60 & +2.88 & 62.42 & 73.14 & +10.78 \\
        
        \bottomrule
        \end{tabular}%
    }
    \caption{Analysis of metric reliability on HalOmi. We compare our training signal (SAC) against testing metrics (Coverage, XCOMET) and word alignment baselines (WSPAlign~\citep{wu2023wspalign}). \textbf{Left:} Average quality scores (for SAC, values denote per-tag accuracy percentage). \textbf{Right:} Pearson correlation with human labels.}
    \label{tab:motivation_analysis}
\end{table*}

\section{Qualitative Analysis}
\label{app:qualitative_analysis}

Table~\ref{tab:case_study} compares the base model (GemmaX2-28-9B), GPT-4o-mini, and M\textsuperscript{2}PO, illustrating how M\textsuperscript{2}PO leverages the \textbf{SAC} signal to rectify partial errors missed by standard rewards (KIWI-XXL).

\paragraph{Mitigating Hallucinations.}
For \textit{En$\to$Zh}, the base model suffers from severe code-switching (``\cn{被疏} evacuated''). Crucially, KIWI-XXL fails to penalize this, assigning a high score (80.08). In contrast, SAC flags it as ``Low Partial''. Guided by this, M\textsuperscript{2}PO eliminates the hallucination to produce a faithful translation (``\cn{被从该地区疏散}''), achieving near-perfect XCOMET (99.98).
\looseness=-1

\paragraph{Capturing Semantic Nuances.}
M\textsuperscript{2}PO outperforms baselines in preserving semantic constraints: \\
\textbf{De$\to$En:} The source ``nur restentleerte'' requires both \textit{only} and \textit{completely} emptied. The base model misses both, while GPT-4o-mini overlooks ``completely''. M\textsuperscript{2}PO captures the full scope (``only submit completely emptied containers''), ensuring maximal fidelity. \\
\textbf{Zh$\to$En:} For ``\cn{口水鸡}'', the base model outputs misleading ``cold chicken'' (High Partial). M\textsuperscript{2}PO generates a precise translation (``Poached chicken with chili sauce''), resolving ambiguity to match the reference.
\looseness=-1

\section{Analysis of Metric Reliability}
\label{app:metric_reliability}

To validate our metric selection, we analyze reliability on the \textit{HalOmi} benchmark~\citep{dale2023halomi}, tracking the performance drop ($\Delta$) when nuanced Partial Errors are included (Table~\ref{tab:motivation_analysis}).

\paragraph{Vulnerability of Standard QE.}
While standard QE models align with general quality, they reveal a severe ``blind spot'' for partial errors. Notably, \textbf{KIWI-XXL} degrades drastically when omissions are included ($\Delta=11.60$), and \textbf{XCOMET} follows a similar trend ($\Delta=10.78$). This indicates a \textit{systematic failure} of regression-based QE to penalize superficially fluent but incomplete translations, necessitating a more semantics-aware signal.

\paragraph{Limitations of Traditional Alignment.}
We explicitly compare against the continuous baseline \textbf{WSPAlign}. Despite capturing lexical correspondences, its correlation with human judgment lags significantly behind LLM-based signals (e.g., $61.97$ vs. $73.99$ for SAC in hallucination). This suggests that traditional alignment tools lack the \textit{deep semantic reasoning} required to identify subtle semantic deviations.

\paragraph{Robustness of LLM-based Signals.}
In contrast, LLM-driven metrics demonstrate superior robustness. Specifically for hallucination evaluation, \textbf{Coverage} achieves the highest correlation ($76.65$) with negligible sensitivity to partial errors ($\Delta=0.47$), validating it as our rigorous testing metric. Crucially, our proposed \textbf{SAC} (used for training) also maintains competitive reliability in detecting hallucinations ($73.99$), confirming that \textit{discrete semantic diagnosis} provides a more accurate and stable supervision signal than traditional continuous baselines.

\section{Implementation Details: Prompts, SAC Configuration, and Costs}
\label{app:implementation_details}

This section details the prompts used for data synthesis and evaluation, the configuration of the Semantic Alignment Classifier (SAC), and the cost analysis for dataset construction and evaluation.

\subsection{Prompt Templates}
\label{subsec:prompts}
\begin{figure*}[t]
    \centering
    \includegraphics[width=0.9\textwidth]{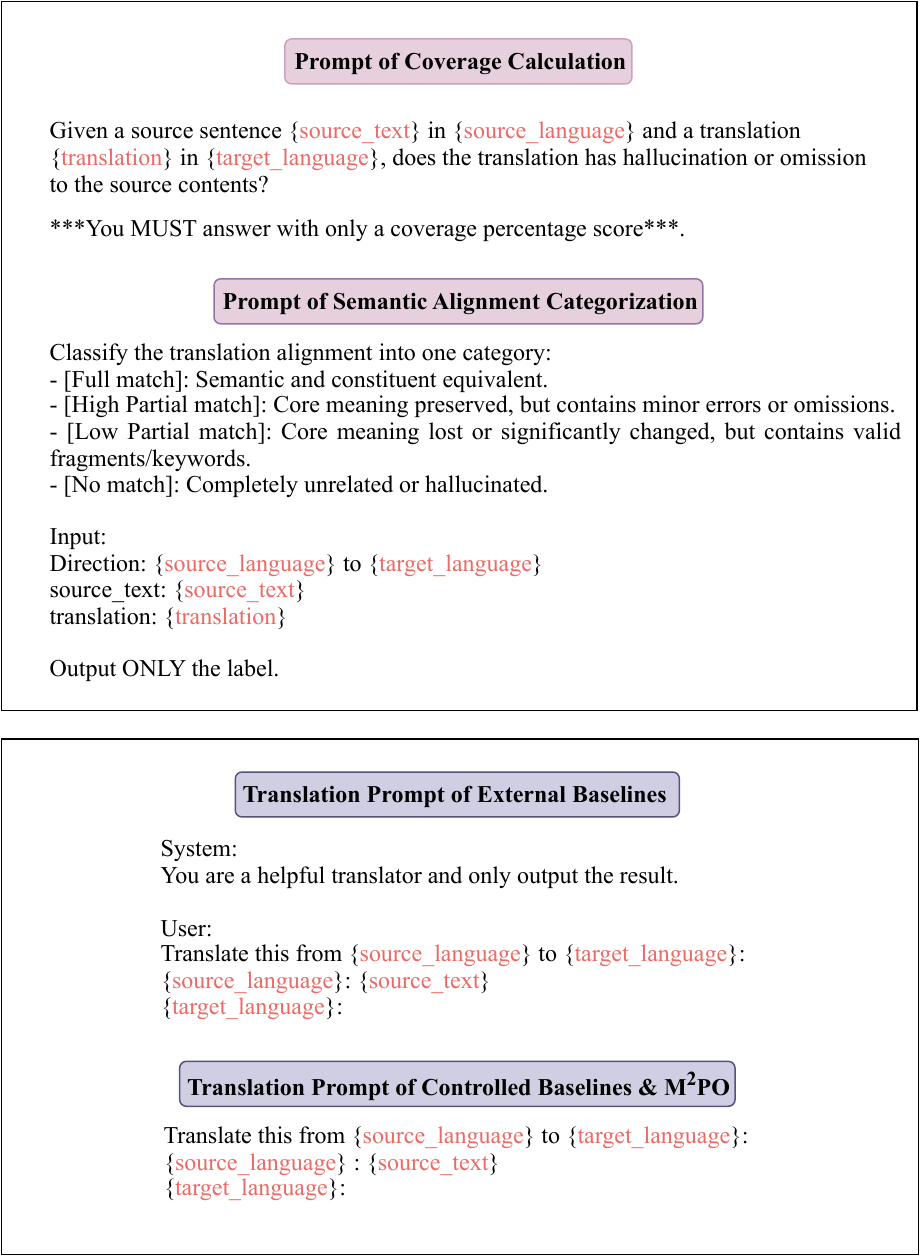} 
    \caption{\textbf{Overview of prompt templates.} 
    \textbf{Top:} Prompts for faithfulness assessment. The Coverage Calculation serves as the primary metric for \textit{test evaluation}, while the Semantic Alignment Categorization (SAC) is employed for constructing preference pairs.
    \textbf{Bottom:} The inference prompts for translation tasks, distinguishing between the format for \textit{External Baselines} and the native format for our \textit{Base Model}, \textit{Controlled Baselines}, and \textit{M\textsuperscript{2}PO}.}
    \label{fig:model_prompts}
\end{figure*}

Figure~\ref{fig:model_prompts} presents the prompt templates used in our study. For \textbf{faithfulness assessment}, we employ the \textit{Coverage Calculation} prompt to query hallucination/omission scores and the \textit{SAC} prompt for discrete faithfulness labeling. For \textbf{translation}, to isolate algorithmic gains, we enforce a strict control strategy where external baselines (proprietary and open-source) follow default instructions, while M\textsuperscript{2}PO and controlled baselines share an identical native task format (based on GemmaX2-28-9B). This translation prompt remains invariant across sampling, training, and inference, ensuring all improvements are driven by our alignment algorithm rather than prompt engineering.

\subsection{SAC Formulation and Sensitivity Analysis}
\label{app:sac_details}
To compute the alignment score $S_{align}$ (Eq.~\ref{eq:static_score_multiplicative}), we utilize GPT-4o-mini. Beyond its cost-efficiency and instruction-following capabilities, this choice strategically decouples training supervision from evaluation, as our final faithfulness benchmarking relies on Gemini-2.0-Flash. This separation mitigates the risk of the model ``gaming'' a specific evaluator's biases. Specifically, the model predicts a semantic category $c$, which is subsequently mapped to a discrete scalar via the following schedule:
\begin{equation}
\resizebox{0.9\linewidth}{!}{$
    \displaystyle
    S_{\text{align}} = \mathcal{M}(c) = 
    \begin{cases} 
    1.0 & \text{if } c \text{ is \textit{Full Match}} \\
    0.7 & \text{if } c \text{ is \textit{High Partial}} \\
    0.3 & \text{if } c \text{ is \textit{Low Partial}} \\
    0.1 & \text{if } c \text{ is \textit{No Match}}
    \end{cases}
$}
\label{eq:score_mapping}
\end{equation}

\paragraph{Sensitivity Analysis.}
To validate our heuristic coefficients, we benchmark our \textit{Stepwise} strategy against three variants: Strict Binary (zero-tolerance), Relaxed Penalty (milder sanctions), and a Zero Floor baseline.
As shown in Table~\ref{tab:sac_sensitivity}, \textit{Strict Binary} yields high Coverage (97.90) but degrades quality (89.02 XCOMET) by discarding useful partial signals. Conversely, \textit{Relaxed Penalty} fails to sufficiently curb hallucinations.
Crucially, the \textit{Zero Floor} setting, which strictly assigns zero weight to severe errors, underperforms our approach. This confirms that a non-zero lower bound acts as a vital \textit{safety floor}, providing robustness against potential classifier noise while maintaining gradient flow.
Ultimately, our strategy achieves the optimal balance, attaining the highest XCOMET (89.67) with near-top faithfulness (97.87).

\subsection{Cost Analysis}
\label{subsec:cost_analysis}

We analyze the cost efficiency of our framework to ensure reproducibility and accessibility. As detailed in Table~\ref{tab:cost_analysis}, the construction of M\textsuperscript{2}PO-Prefer involved substantial scale, leveraging the cost-effective GPT-4o-mini for $\sim$110k candidate generations and $\sim$160k SAC assessments. The latter required processing a significant token volume ($\sim$57M) to handle the detailed instruction definitions needed for accurate alignment categorization.

Furthermore, our evaluation phase utilized Gemini-2.0-Flash to conduct an extensive benchmarking campaign. We computed faithfulness (Coverage) scores for approx. 15 comparative systems and ablation variants across all test sets, totaling $\sim$320k assessment requests. 
Remarkably, by strategically prioritizing these high-efficiency models, the cumulative API expenditure for the entire project remained negligible (\textbf{$\approx$\$20 USD}\footnote{Pricing estimates are based on official documentation as of December 2025: OpenAI (\url{https://platform.openai.com/docs/pricing}) and Google (\url{https://ai.google.dev/gemini-api/docs/pricing}).}). We leverage the ultra-low pricing of GPT-4o-mini (\$0.15/\$0.60 per 1M input/output tokens) for reliable construction and Gemini-2.0-Flash (\$0.10/\$0.40 per 1M input/output tokens) for large-scale evaluation.
This demonstrates that M\textsuperscript{2}PO is highly accessible to the research community, enabling rigorous, large-scale alignment research without requiring prohibitive budgets.

\begin{table}[t]
    \centering
    \small
    \setlength{\tabcolsep}{1.5pt}
    \renewcommand{\arraystretch}{1.3} 
    \begin{tabular}{l c c c}
    \toprule
    \textbf{Configuration} & \textbf{Weights} & \textbf{Coverage} & \textbf{XCOMET} \\
    \midrule
    \textit{Strict Binary} & $\{1.0, 0.0, 0.0, 0.0\}$ & \textbf{97.90} & 89.02 \\
    \textit{Relaxed Penalty} & $\{1.0, 0.9, 0.5, 0.1\}$ & 97.53 & 89.55 \\
    \textit{Zero Floor} & $\{1.0, 0.7, 0.3, 0.0\}$ & 97.71 & 89.58 \\
    \rowcolor{gray!10} \textbf{Ours (Stepwise)} & $\{1.0, 0.7, 0.3, 0.1\}$ & 97.87 & \textbf{89.67} \\
    \bottomrule
    \end{tabular}
    \caption{Sensitivity analysis of SAC weight configurations on the WMT23 benchmark.}
    \label{tab:sac_sensitivity}
\end{table}

\begin{table}[t]
    \centering
    \resizebox{\linewidth}{!}{%
        \setlength{\tabcolsep}{6pt}
        \renewcommand{\arraystretch}{1.2}
        \begin{tabular}{l cc r}
        \toprule
        \multirow{2}{*}{\textbf{Stage}} & \multicolumn{2}{c}{\textbf{Usage Statistics}} & \textbf{Expenditure} \\
        \cmidrule(lr){2-3} \cmidrule(lr){4-4}
         & \textbf{Volume} & \textbf{Tokens} & \textbf{Cost (USD)} \\
        \midrule
        
        \rowcolor{gray!15} 
        \multicolumn{4}{l}{\textit{Construction (GPT-4o-mini)}} \\
        Candidate Gen. & 110k & $\sim$16.5M & 4.29 \\
        SAC Assessment\textdagger & 160k & $\sim$56.8M & 8.90 \\
        \midrule
        
        \rowcolor{gray!15}
        \multicolumn{4}{l}{\textit{Evaluation (Gemini-2.0-Flash)}} \\
        Faithfulness Eval.\textdaggerdbl & 320k & $\sim$65.6M & 7.04 \\
        \midrule
        
        \textbf{Total Project} & - & $\sim$138.9M & \textbf{$\approx$ 20.23} \\
        \bottomrule
        \end{tabular}%
    }
\caption{Cost breakdown of the M\textsuperscript{2}PO framework. Expenditures are calculated based on Dec 2025 pricing for GPT-4o-mini (\$0.15/\$0.60 per 1M tokens for input/output) and Gemini-2.0-Flash (\$0.10/\$0.40). \textdagger: High token usage reflects detailed instruction prompts. \textdaggerdbl: Covers extensive evaluation of approx. 15 comparative systems across all benchmarks.}
    \label{tab:cost_analysis}
\end{table}

\section{Dataset Statistics and Scaling Analysis}
\label{app:dataset_statistics}

\paragraph{Dataset Construction.}
Table~\ref{tab:dataset_statistics} summarizes the raw seed corpora ($\sim$28,000 source sentences), which are strictly isolated from testing benchmarks to ensure zero data leakage. To construct the final M\textsuperscript{2}PO-Prefer dataset, we expand these seeds via hybrid generation ($K=8$) and apply rigorous curation to maximize signal clarity. Specifically, we perform (1) \textit{deduplication} of redundant model outputs and (2) \textit{margin filtering} based on QE scores to exclude ambiguous pairs with insufficient quality separation ($\Delta_{\text{qe}} < 2$). This pipeline yields a refined set of 20,000 high-quality instances (comprising 160,000 total candidates), prioritizing discriminative power over raw volume.

\begin{figure}[t]
  \centering
  \includegraphics[width=\linewidth]{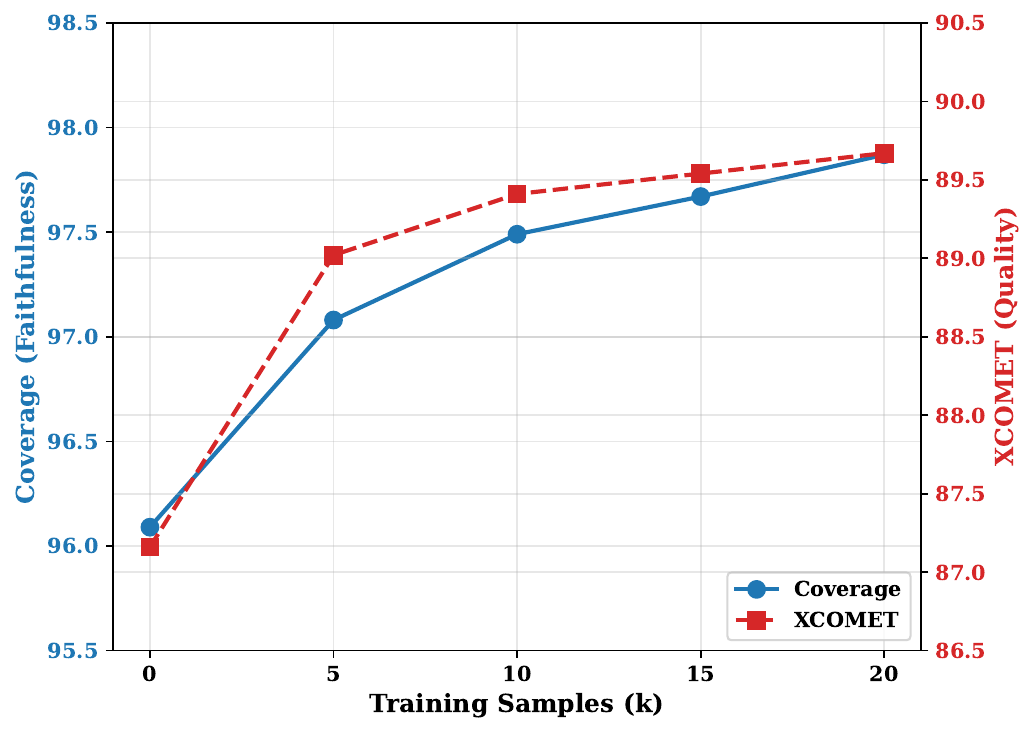}
  \caption{Scaling analysis of M\textsuperscript{2}PO on varying proportions of the training set.}
  \label{fig:data_scaling}
\end{figure}

\paragraph{Scaling Analysis.}
To validate the sufficiency of our curated data scale, we investigate the relationship between training volume and performance. Figure~\ref{fig:data_scaling} illustrates the scaling curves for Coverage and XCOMET as the proportion of M\textsuperscript{2}PO-Prefer increases from 0\% to 100\%. 

The results underscore remarkable data efficiency: utilizing merely 25\% of the refined set ($\sim$5,000 samples) leads to a sharp performance leap, achieving a +1.86 gain in XCOMET and +0.99 in Coverage. This initial surge captures more than half of the total improvement observed at full scale. Beyond this point, the performance follows a logarithmic scaling pattern with diminishing marginal returns. The highly synchronized growth of both metrics confirms that the multi-pair objective effectively extracts robust alignment signals even from a limited pool of high-quality preference pairs, justifying our focus on data quality during construction.

\begin{table}[t]
    \centering
    \resizebox{\columnwidth}{!}{%
        \begin{tabular}{llcccc}
        \toprule
        & \textbf{Dataset} & \textbf{Lang} & \textbf{$|S|$} & \textbf{$|T|$} & \textbf{$|L|$} \\
        \midrule
        \multirow{9}{*}{\rotatebox[origin=c]{90}{\em Training}} 
        & \multirow{3}{*}{WMT22} 
          & En$\leftrightarrow$Zh & 3.9K & 82.4K & 21 \\
        & & En$\leftrightarrow$De & 4.0K & 65.9K & 16 \\
        & & En$\leftrightarrow$Ja & 4.0K & 63.1K & 16 \\
        \cline{2-6}\noalign{\vskip 0.5ex}
        & \multirow{3}{*}{IWSLT 15--17} 
          & En$\leftrightarrow$Zh & 2.2K & 35.7K & 16 \\
        & & En$\leftrightarrow$De & 4.0K & 67.7K & 17 \\
        & & En$\leftrightarrow$Ja & 4.0K & 67.2K & 17 \\
        \cline{2-6}\noalign{\vskip 0.5ex}
        & \multirow{3}{*}{FLORES-200-dev} 
          & En$\leftrightarrow$Zh & 2.0K & 41.9K & 21 \\
        & & En$\leftrightarrow$De & 2.0K & 41.9K & 21 \\
        & & En$\leftrightarrow$Ja & 2.0K & 41.9K & 21 \\
        \midrule
        \multirow{9}{*}{\rotatebox[origin=c]{90}{\em Testing}}
        & \multirow{3}{*}{WMT23} 
          & En$\leftrightarrow$Zh & 4.1K & 82.8K & 20 \\
        & & En$\leftrightarrow$De & 1.1K & 60.6K & 55 \\
        & & En$\leftrightarrow$Ja & 4.1K & 68.6K & 17 \\
        \cline{2-6}\noalign{\vskip 0.5ex}
        & \multirow{3}{*}{FLORES-200-test} 
          & En$\leftrightarrow$Zh & 2.0K & 43.8K & 22 \\
        & & En$\leftrightarrow$De & 2.0K & 43.8K & 22 \\
        & & En$\leftrightarrow$Ja & 2.0K & 43.8K & 22 \\
        \cline{2-6}\noalign{\vskip 0.5ex}
        & \multirow{3}{*}{WMT24} 
          & En$\rightarrow$Zh & 1.0K & 32.3K & 32 \\
        & & En$\rightarrow$De & 2.0K & 64.7K & 32 \\
        & & En$\rightarrow$Ja & 2.0K & 64.7K & 32 \\
        \bottomrule
        \end{tabular}%
    }
  \caption{Statistics of the datasets employed in this study. We distinguish between \textit{Training Sources} (seed data for preference construction) and disjoint \textit{Testing Benchmarks}, which cover General (\textsc{Wmt}), Spoken (\textsc{Iwslt}), and Encyclopedic (\textsc{Flores}) domains. Reported metrics include sentence count ($|S|$), English token count ($|T|$), and average sentence length ($|L|$).}
    \label{tab:dataset_statistics}
\end{table}

\begin{table}[t]
    \centering
    \resizebox{\linewidth}{!}{%
        \renewcommand{\arraystretch}{1.2}
        \begin{tabular}{lcc}
        \toprule
        \textbf{Model / Method} & \textbf{Coverage} & \textbf{XCOMET} \\
        \midrule
        \textit{Base (GemmaX2-28-9B)} & 96.09 & 87.16 \\
        \midrule
        \multicolumn{3}{l}{\textit{SFT Baselines}} \\
        \quad + M\textsuperscript{2}PO-Prefer (Top-ranked) & 96.81 & 87.79 \\
        \quad + Gold Reference & 97.06 & 87.72 \\
        \midrule
        \multicolumn{3}{l}{\textit{Preference Optimization (Ours)}} \\
        \rowcolor{gray!15} 
        \quad + \textbf{M\textsuperscript{2}PO} & \textbf{97.87} & \textbf{89.67} \\
        \bottomrule
        \end{tabular}%
    }
    \caption{Comparison of SFT baselines trained on different targets versus M\textsuperscript{2}PO.}
    \label{tab:sft_targets_comparison}
\end{table}

\section{Extended Ablation on SFT Target Construction}
\label{app:baseline_ablations}

To determine whether our Supervised Fine-Tuning (SFT) baseline is bottlenecked by using top-ranked model candidates, we evaluate an alternative SFT model trained strictly on authentic \textit{Gold References}. As Table~\ref{tab:sft_targets_comparison} shows, while the Gold Reference naturally yields higher faithfulness (Coverage: 97.06 vs. 96.81 for top-ranked), M\textsuperscript{2}PO still consistently outperforms this optimal SFT setup. This gap highlights a fundamental limitation of Maximum Likelihood Estimation (MLE): even given perfect human references, standard SFT merely mimics the target distribution. It lacks the discriminative capability to actively penalize fine-grained partial errors (e.g., subtle omissions) that our multi-pair preference objective effectively mitigates.

\section{Additional Experimental Results}
\label{app:additional_experiments}

To rigorously verify the robustness and generalization capabilities of M\textsuperscript{2}PO, we extend our evaluation beyond the primary benchmark. We incorporate \textbf{FLORES-200-test} to assess domain transfer performance on diverse topics and \textbf{WMT24} to evaluate temporal robustness against newer data distributions. Additionally, we provide supplementary reference-based evaluations using \textbf{COMET-22} across all three datasets to ensure metric consistency.

As detailed in Tables~\ref{tab:wmt23_comet22}, \ref{tab:flores_coverage_xcomet}, \ref{tab:flores200_comet22}, and \ref{tab:wmt24_results}, M\textsuperscript{2}PO consistently outperforms strong baselines across these varied settings. 
The COMET-22 results on WMT23 strongly corroborate the quality gains observed in the main text. Furthermore, the strong performance on FLORES-200 confirms that our faithfulness constraints transfer effectively to out-of-distribution domains, while the WMT24 results demonstrate the model's resilience to temporal shifts. These findings collectively indicate that M\textsuperscript{2}PO learns generalized preference patterns rather than overfitting to specific training data or evaluation metrics.

\definecolor{effgreen}{RGB}{46, 139, 87}  
\definecolor{effred}{RGB}{178, 34, 34}    
\newcommand{\fast}{{\textbf{1$\times$}}}
\newcommand{\slow}{{8$\times$}}

\begin{table*}[t]
    \centering
    \small
    \renewcommand{\tabularxcolumn}[1]{>{\raggedright\arraybackslash}m{#1}} 
    \renewcommand{\arraystretch}{1.25} 
    \setlength{\tabcolsep}{8pt} 
    
    \definecolor{sac-high}{HTML}{D9730D}
    \definecolor{sac-full}{HTML}{0F7B6C}
    \newcommand{\unitfont}[1]{{\scriptsize #1}}

    \begin{tabularx}{\textwidth}{l X c c c} 
    \toprule
    \textbf{System} & \textbf{Translation Output} & 
    \makecell{\unitfont{\textbf{MetricX-24-XXL}}\\\unitfont{($\downarrow$)}} & 
    \makecell{\unitfont{\textbf{KIWI-XXL}}\\\unitfont{($\uparrow$)}} & 
    \unitfont{\textbf{SAC}} \\
    \midrule
    
    \multicolumn{5}{l}{\cellcolor{gray!15}\textbf{Source (de$\to$en)}: Entsorgung: Bitte \textcolor{blue}{nur restentleerte} Gebinde dem Recycling zuführen.} \\
    \midrule
    GemmaX2-28-9B & Disposal: Please submit \textcolor{highpartial}{emptied} containers for recycling. & 4.03 & 69.38 & \textbf{\textcolor{sac-high}{High Partial}} \\
    GPT-4o-mini & Disposal: Please \textcolor{highpartial}{only} contribute \textcolor{highpartial}{emptied} containers to recycling. & 5.25 & 45.48 & \textbf{\textcolor{sac-full}{Full match}} \\
    M\textsuperscript{2}PO (Ours) & Disposal: Please \textcolor{fullmatch}{only} submit \textcolor{fullmatch}{completely emptied} containers to recycling. & 3.39 & 76.71 & \textbf{\textcolor{sac-full}{Full match}} \\
    \bottomrule
    \end{tabularx}
    \caption{Qualitative comparison of continuous reward proxies (MetricX-24-XXL and KIWI-XXL) and SAC on translations containing partial errors.}
    \label{tab:metric_case_analysis}
\end{table*}

\begin{table}[t]
    \centering
    \footnotesize
    \resizebox{\linewidth}{!}{%
        \renewcommand{\arraystretch}{1.3}
        \begin{tabular}{ll cc c}
        \toprule
        \textbf{Method} & \textbf{Reward Signal} & \textbf{Coverage} & \textbf{XCOMET} & \textbf{Cost} \\
        \midrule
        \textit{Base} & N/A & 96.09 & 87.16 & - \\
        \midrule
        \rowcolor{gray!15} 
        \multicolumn{5}{l}{\textit{Online-RL (Group Size = 8)}} \\
        GRPO & KIWI-XXL only & 97.14 & 89.01 & 8$\times$ \\
        GRPO & Augmented Reward & 97.55 & 89.15 & 8$\times$ \\
        GSPO & KIWI-XXL only & 97.35 & 88.94 & 8$\times$ \\
        GSPO & Augmented Reward & 97.76 & 89.28 & 8$\times$ \\
        \midrule
        \rowcolor{gray!15} 
        \multicolumn{5}{l}{\textit{Offline (Ours)}} \\
        \textbf{M\textsuperscript{2}PO} & Augmented Reward & \textbf{97.87} & \textbf{89.67} & \textbf{1$\times$} \\
        \bottomrule
        \end{tabular}%
    }
    \caption{Fair comparison between M\textsuperscript{2}PO and Online-RL methods. All methods in the comparison use the same composite reward signal to isolate the impact of the optimization framework. \textit{Cost} denotes relative training time.}
    \label{tab:online_rl_comparison}
\end{table}

\section{Comparison with Online-RL Approaches}
\label{app:online_rl}

We benchmark M\textsuperscript{2}PO against prominent Online-RL optimizers, specifically GRPO~\citep{guo2025deepseek} and GSPO~\citep{zheng2025group}, evaluating both empirical performance and underlying algorithmic mechanisms.

\paragraph{Empirical Superiority: Quality and Efficiency.} 
To isolate algorithmic gains from reward quality, we align the reward proxies across all baselines. As shown in Table~\ref{tab:online_rl_comparison}, while the \textbf{Augmented Reward} (KIWI-XXL+SAC+Model Confidence) enhances all models, M\textsuperscript{2}PO consistently maintains a superior performance frontier. Under identical reward conditions, it yields improvements ranging from 0.11 to 0.32 in Coverage and 0.39 to 0.52 in XCOMET. Beyond quality gains, M\textsuperscript{2}PO bypasses the online sampling bottleneck and the complexity of group-wide advantage calculations, achieving an 8$\times$ speedup in GPU hours compared to Online-RL baselines. This dual advantage in both generation fidelity and computational cost establishes M\textsuperscript{2}PO as a highly scalable and effective alignment paradigm.

\paragraph{Mechanistic Analysis: Structured Contrast vs. Group Mean.} 
The performance gap between M\textsuperscript{2}PO and group-based counterparts highlights a fundamental difference in how these algorithms utilize candidate sets. Group-based methods (GRPO/GSPO) treat the candidate pool as an unstructured set, optimizing each candidate relative to a uniform group average ($r_i - \bar{r}$). While effective for general alignment, this aggregation inherently dilutes the specific, fine-grained contrast between a high-quality translation and a targeted ``hard negative'' (e.g., a fluent but partially unfaithful candidate). In contrast, M\textsuperscript{2}PO is explicitly designed to exploit the ordered, hierarchical structure of the candidate list. Rather than evaluating against an aggregated baseline, M\textsuperscript{2}PO's pairwise coupling isolates the exact semantic boundaries the model needs to learn. By structurally enforcing a direct contrast between specific preferred and rejected pairs, it extracts a more precise and targeted learning signal for subtle translation errors than a generalized group mean can provide.

\section{Qualitative Analysis of Metric Reliability}
\label{app:metric_comparison}

To address whether advanced regression metrics like MetricX-24-XXL can eliminate the need for explicit faithfulness supervision, Table~\ref{tab:metric_case_analysis} illustrates their limitations when evaluating partial errors.

\paragraph{The Blind Spot of Continuous Metrics.} 
Continuous metrics often exhibit a critical blind spot: prioritizing surface-level fluency over strict semantic faithfulness. In the \textit{de$\to$en} case, GemmaX2-28-9B commits a \textit{Major Omission} (missing ``nur'' and ``restentleerte''). Yet, it receives a ``better'' distance score from MetricX-24-XXL (4.03) than GPT-4o-mini (5.25), which only has a \textit{Minor Omission}. KIWI-XXL shows a similar ranking inversion (69.38 vs. 45.48). This confirms that even leading metrics like MetricX-24-XXL fail to reliably penalize unfaithful translations if the output remains highly fluent.

\paragraph{The Necessity of SAC.} 
Relying solely on such metrics provides misleading training signals that favor fluent omissions. Our SAC classifier acts as an essential semantic guardrail. As shown, it accurately detects the severe semantic loss in the GemmaX2-28-9B output (assigning a High Partial penalty) while correctly recognizing GPT-4o-mini as a Full match. By integrating this discrete supervision, our framework successfully overrides the ranking biases of continuous metrics, ensuring optimization is strictly anchored to semantic faithfulness.

\begin{table}[t]
    \centering
    \small
    \renewcommand{\arraystretch}{1.4}
    \setlength{\tabcolsep}{10pt}
    \begin{tabular}{l cc}
    \toprule
    \textbf{$\lambda_{\text{rank}}$ Value} & \textbf{Coverage} & \textbf{XCOMET} \\
    \midrule
    0.0 \textit{(w/o $\mathcal{L}_{\text{Rank}}$)} & 97.15 & 89.19 \\
    0.1 & 97.58 & 89.45 \\
    \rowcolor{gray!15} \textbf{0.5 (Ours)} & \textbf{97.87} & \textbf{89.67} \\
    1.0 & 97.63 & 89.51 \\
    2.0 & 97.28 & 89.24 \\
    \bottomrule
    \end{tabular}
    \caption{Sensitivity analysis of the ranking loss weight $\lambda_{\text{rank}}$ on the WMT23 benchmark.}
    \label{tab:lambda_sensitivity}
\end{table}

\section{Impact of Ranking Regularization Weight (\texorpdfstring{$\lambda_{\text{rank}}$}{lambda rank})}
\label{app:lambda_rank_analysis}

We conduct a sensitivity analysis on the ranking regularization weight $\lambda_{\text{rank}}$ to determine its optimal contribution to the joint objective. Table~\ref{tab:lambda_sensitivity} summarizes the performance on WMT23 across a value spectrum of $\{0.0, 0.1, 0.5, 1.0, 2.0\}$.

The results exhibit a clear inverted U-shaped trend. Starting from the baseline without ranking regularization ($\lambda_{\text{rank}}=0.0$), we observe steady improvements as the weight increases. Specifically, the configuration with $\lambda_{\text{rank}}=0.5$ reaches the performance peak, achieving significant gains of +0.72 in Coverage and +0.48 in XCOMET compared to the baseline. This confirms that incorporating global ordinal consistency effectively complements the local pairwise optimization, helping the model distinguish fine-grained quality differences.

However, the performance begins to degrade when the weight exceeds this threshold ($\lambda_{\text{rank}} \ge 1.0$). This decline suggests that an overly aggressive ranking penalty creates a soft constraint that competes with the primary CPO objective. When the ranking loss dominates, it forces the model to rigidly fit the target distribution rather than focusing on maximizing the decision margin for the optimal translation. Consequently, we adopt $\lambda_{\text{rank}} = 0.5$ as the default setting to balance global consistency with local discriminative power.

\section{Impact of Dynamic Weighting Strategy (\texorpdfstring{$\alpha_t$}{alpha t})}
\label{app:static_vs_dynamic}

To investigate the sensitivity of our dynamic curriculum ($\alpha_t$) to boundary constraints, we compare the proposed schedule against fixed and alternative dynamic ranges. Table~\ref{tab:alpha_ablation} confirms that while dynamic scheduling generally outperforms fixed baselines, the choice of boundary values is pivotal. The \textit{Narrow Schedule} ($0.3 \to 0.7$) underperforms due to insufficient decoupling of external anchoring and internal refinement. While the \textit{Full Range} ($0.0 \to 1.0$) enhances quality, it suffers from faithfulness regression driven by ``self-delusion''—unchecked hallucinations at $\alpha_t=1.0$. Consequently, our \textit{Clipped Schedule} ($0.1 \to 0.9$) strikes the optimal balance; retaining a 10\% \textit{safety anchor} ensures semantic fidelity while maximizing self-refinement.

\begin{table}[t]
    \centering
    \small
    \setlength{\aboverulesep}{0pt}
    \setlength{\belowrulesep}{0pt}
    
    \setlength{\tabcolsep}{4.5pt} 
    \renewcommand{\arraystretch}{1.25} 
    
    \resizebox{\columnwidth}{!}{%
    \begin{tabular}{l cc}
        \toprule
        \multirow{2}{*}{\textbf{Configuration}} & \multicolumn{2}{c}{\textbf{Performance}} \\
        \cmidrule(lr){2-3}
         & \textbf{Coverage} & \textbf{XCOMET} \\
        \midrule
        \textit{Base (GemmaX2-28-9B)} & 96.09 & 87.16 \\
        \midrule
        
        \rowcolor{gray!15} 
        \multicolumn{3}{l}{\textit{Fixed Strategies (Static Weight)}} \\
        \hspace{1em} $\alpha \equiv 0.0$ (Static Only) & 96.98 & 89.44 \\
        \hspace{1em} $\alpha \equiv 0.5$ (Equal Mix)   & 97.01 & 89.08 \\
        \hspace{1em} $\alpha \equiv 1.0$ (Conf. Only)  & 96.67 & 88.32 \\
        \midrule
        
        \rowcolor{gray!15} 
        \multicolumn{3}{l}{\textit{Dynamic Schedules ($\alpha_t: \text{Start} \to \text{End}$)}} \\
        \hspace{1em} $0.3\to0.7$ (Narrow) & 97.35 & 89.48 \\
        \hspace{1em} $0.0 \to 1.0$ (Full Range) & 97.62 & 89.55 \\
        \hspace{1em} \textbf{$0.1 \to 0.9$ (Clipped)} & \textbf{97.87} & \textbf{89.67} \\
        \bottomrule
    \end{tabular}%
    }
    \caption{Ablation study on the curriculum schedule $\alpha_t$ on the WMT23. We compare fixed scalar weights against dynamic schedules with varying boundaries.}
    \label{tab:alpha_ablation}
\end{table}

\clearpage

\begin{table*}[t]
\centering
\scriptsize
\resizebox{\textwidth}{!}{%
\setlength{\tabcolsep}{8pt}
\begin{tabular}{lccccccc}
\toprule
\textbf{Model} & \textbf{AVG} & \textbf{En$\to$Zh} & \textbf{En$\to$De} & \textbf{En$\to$Ja} & \textbf{Zh$\to$En} & \textbf{De$\to$En} & \textbf{Ja$\to$En} \\
\midrule
\cellcolor{c_closed}Gemini-2.0-Flash & 84.82 & 86.20 & 84.50 & \bb{88.11} & 80.87 & 85.86 & \bb{83.37} \\
\cellcolor{c_closed}GPT-4o & \bb{84.92} & 86.57 & \bb{84.53} & 87.68 & 81.33 & \bb{86.17} & 83.23 \\
\cellcolor{c_closed}GPT-4o-mini & 84.66 & 86.33 & 84.15 & 87.48 & 81.31 & 85.70 & 82.96 \\
\midrule

\cellcolor{c_open}Aya-expanse-32B & 84.11 & 86.31 & 82.10 & 87.70 & 80.62 & 85.22 & 82.69 \\
\cellcolor{c_open}Aya-23-35B & 82.81 & 84.74 & 80.66 & 86.44 & 79.37 & 84.15 & 81.49 \\
\cellcolor{c_open}TowerInstruct-13B & 80.84 & 84.98 & 79.28 & 77.08 & 80.16 & 84.38 & 79.17 \\
\cellcolor{c_open}ALMA-13B-R & 81.98 & 83.99 & 79.88 & 83.77 & 80.42 & 84.57 & 79.27 \\
\midrule

\cellcolor{c_ours}Qwen3-4B-Instruct & 81.42 & 84.10 & 76.88 & 84.27 & 80.41 & 82.57 & 80.27 \\
\cellcolor{c_ours}\hspace{1em}+ SFT & 82.24 & 85.28 & 77.13 & 84.96 & 80.88 & 84.08 & 81.13 \\
\cellcolor{c_ours}\hspace{1em}+ CPO & 82.78 & 86.39 & 77.47 & 86.05 & 81.00 & 84.11 & 81.66 \\
\cellcolor{c_ours}\hspace{1em}\textbf{+ M\textsuperscript{2}PO} & 83.07 & 86.47 & 78.19 & 86.37 & 81.16 & 84.35 & 81.87 \\
\hline

\cellcolor{c_ours}GemmaX2-28-9B & 83.36 & 85.99 & 81.59 & 85.91 & 80.52 & 84.73 & 81.43 \\
\cellcolor{c_ours}\hspace{1em}+ SFT & 83.74 & 86.64 & 81.77 & 86.31 & 80.87 & 84.94 & 81.93 \\
\cellcolor{c_ours}\hspace{1em}+ CPO & 84.03 & 86.73 & 81.98 & 86.82 & 81.18 & 85.11 & 82.37 \\
\cellcolor{c_ours}\hspace{1em}\textbf{+ M\textsuperscript{2}PO} & \cc{84.61} & \bbcc{86.93} & \cc{82.34} & \cc{88.06} & \bbcc{81.63} & \cc{85.42} & \cc{83.26} \\
\bottomrule
\end{tabular}%
}
\caption{\setlength{\fboxsep}{1pt}
Main results on the WMT23 benchmark. Results are reported in \textbf{COMET22} scores. Model names are color-coded by type: \colorbox{c_closed}{Proprietary}, \colorbox{c_open}{Open-source baselines}, and \colorbox{c_ours}{Our experiments}. \textbf{Bold} indicates the best result overall. \colorbox{gray!20}{Colored background} indicates the best result among open-source models.}
\label{tab:wmt23_comet22}
\end{table*}

\begin{table*}[t]
\centering
\footnotesize
\resizebox{\textwidth}{!}{%
\setlength{\tabcolsep}{3.5pt}
\renewcommand{\arraystretch}{1.15}
\begin{tabular}{lccccccc}
\toprule
\textbf{Model} & \textbf{AVG} & \textbf{En$\to$Zh} & \textbf{En$\to$De} & \textbf{En$\to$Ja} & \textbf{Zh$\to$En} & \textbf{De$\to$En} & \textbf{Ja$\to$En} \\
\midrule
\cellcolor{c_closed}Gemini-2.0-Flash & 98.04 / 95.10 & 97.29 / 91.97 & 99.05 / 98.01 & 97.74 / 93.29 & 98.20 / 96.49 & 98.79 / 96.27 & 97.17 / 94.56 \\
\cellcolor{c_closed}GPT-4o & \bb{98.42} / 95.21 & 97.62 / 91.20 & \textbf{99.43} / 98.18 & 98.39 / 93.43 & 98.58 / 96.73 & \textbf{98.96} / 96.64 & \textbf{97.55} / \textbf{95.06} \\
\cellcolor{c_closed}GPT-4o-mini & 98.30 / 94.66 & 97.38 / 90.01 & 99.37 / 97.74 & 98.26 / 92.95 & 98.56 / 96.54 & 98.84 / 96.22 & 97.36 / 94.50 \\
\midrule

\cellcolor{c_open}Aya-expanse-32B & 97.92 / 95.02 & 97.05 / 91.87 & 99.11 / \bbcc{98.39} & 97.73 / 93.63 & 98.11 / 96.64 & 98.55 / 96.10 & 96.98 / 93.48 \\
\cellcolor{c_open}Aya-23-35B & 97.46 / 94.10 & 96.67 / 90.90 & 98.87 / 98.15 & 97.06 / 92.30 & 97.59 / 96.16 & 98.28 / 94.99 & 96.30 / 92.08 \\
\cellcolor{c_open}TowerInstruct-13B & 97.25 / 93.03 & 96.45 / 89.60 & 99.12 / 98.05 & 95.87 / 86.16 & 97.54 / 96.14 & 98.32 / 95.94 & 96.19 / 92.39 \\
\cellcolor{c_open}ALMA-13B-R & 96.39 / 92.20 & 94.77 / 89.32 & 98.22 / 98.21 & 94.60 / 81.88 & 97.25 / 96.25 & 98.49 / 96.57 & 94.98 / 90.99 \\
\midrule

\cellcolor{c_ours}Qwen3-4B-Instruct & 96.13 / 92.66 & 96.14 / 89.63 & 95.88 / 96.65 & 96.82 / 90.53 & 97.09 / 94.81 & 96.68 / 94.14 & 94.14 / 90.17 \\
\cellcolor{c_ours}\hspace{1em}+ SFT & 96.64 / 93.86 & 96.53 / 90.91 & 97.15 / 97.45 & 96.87 / 91.64 & 97.15 / 95.34 & 97.44 / 95.68 & 94.68 / 92.11 \\
\cellcolor{c_ours}\hspace{1em}+ CPO & 97.15 / 94.14 & 97.01 / 91.28 & 97.05 / 97.82 & 97.26 / 91.93 & 97.46 / 95.82 & 97.82 / 95.52 & 96.32 / 92.49 \\
\cellcolor{c_ours}\hspace{1em}\textbf{+ M\textsuperscript{2}PO} & 98.03 / 94.76 & 97.32 / 91.84 & 98.95 / 98.10 & 97.98 / 92.49 & 98.21 / 96.39 & 98.62 / 95.96 & 97.12 / 93.77 \\
\hline

\cellcolor{c_ours}GemmaX2-28-9B & 96.87 / 93.63 & 96.91 / 90.69 & 98.07 / 97.67 & 96.60 / 90.95 & 97.30 / 95.47 & 97.23 / 95.28 & 95.13 / 91.69 \\
\cellcolor{c_ours}\hspace{1em}+ SFT & 97.66 / 94.68 & 97.20 / 90.69 & 98.71 / 97.53 & 97.26 / 92.85 & 98.15 / 96.64 & 98.12 / 96.37 & 96.53 / 93.99 \\
\cellcolor{c_ours}\hspace{1em}+ CPO & 97.76 / 95.18 & 97.12 / 92.13 & 98.75 / 97.24 & 97.38 / 93.60 & 98.21 / 96.73 & 98.25 / 96.65 & 96.83 / 94.70 \\
\cellcolor{c_ours}\hspace{1em}\textbf{+ M\textsuperscript{2}PO} & \cc{98.41} / \bbcc{95.73} & \bbcc{97.68} / \bbcc{92.51} & \cc{99.37} / 98.23 & \bbcc{98.40} / \bbcc{94.50} & \bbcc{98.76} / \bbcc{97.16} & \cc{98.93} / \bbcc{96.98} & \cc{97.34} / \cc{94.98} \\
\bottomrule
\end{tabular}%
}

\caption{
\setlength{\fboxsep}{1pt}
Main results on the FLORES-200-test benchmark. Results are reported in the format of \textbf{Coverage / XCOMET}. Model names are color-coded by type: \colorbox{c_closed}{Proprietary}, \colorbox{c_open}{Open-source baselines}, and \colorbox{c_ours}{Our experiments}. \textbf{Bold} indicates the best result overall. \colorbox{gray!20}{Colored background} indicates the best result among open-source models.}
\label{tab:flores_coverage_xcomet}
\end{table*}

\begin{table*}[t]
\centering
\footnotesize
\resizebox{\textwidth}{!}{%
\setlength{\tabcolsep}{10pt}
\renewcommand{\arraystretch}{1.15}
\begin{tabular}{lccccccc}
\toprule
\textbf{Model} & \textbf{AVG} & \textbf{En$\to$Zh} & \textbf{En$\to$De} & \textbf{En$\to$Ja} & \textbf{Zh$\to$En} & \textbf{De$\to$En} & \textbf{Ja$\to$En} \\
\midrule
\cellcolor{c_closed}Gemini-2.0-Flash & 89.30 & 89.04 & 88.93 & 91.65 & 87.71 & 89.79 & 88.67 \\
\cellcolor{c_closed}GPT-4o & 89.32 & 88.91 & \bb{88.98} & 91.76 & 87.78 & \bb{89.92} & 88.55 \\
\cellcolor{c_closed}GPT-4o-mini & 88.88 & 88.31 & 88.42 & 91.36 & 87.32 & 89.69 & 88.20 \\
\midrule

\cellcolor{c_open}Aya-expanse-32B & 89.22 & 88.85 & 88.48 & 91.64 & \bbcc{87.80} & \cc{89.83} & \bbcc{88.73} \\
\cellcolor{c_open}Aya-23-35B & 88.60 & 87.73 & 87.91 & 90.96 & 87.44 & 89.42 & 88.11 \\
\cellcolor{c_open}TowerInstruct-13B & 88.40 & 88.40 & 88.23 & 89.04 & 87.33 & 89.64 & 87.75 \\
\cellcolor{c_open}ALMA-13B-R & 87.80 & 87.07 & 87.96 & 88.37 & 87.04 & 89.50 & 86.84 \\
\midrule

\cellcolor{c_ours}Qwen3-4B-Instruct & 87.40 & 86.73 & 87.08 & 88.77 & 86.72 & 88.15 & 86.94 \\
\cellcolor{c_ours}\hspace{1em}+ SFT & 88.06 & 87.38 & 87.00 & 90.13 & 87.25 & 89.08 & 87.53 \\
\cellcolor{c_ours}\hspace{1em}+ CPO & 88.52 & 87.67 & 87.90 & 90.23 & 87.28 & 89.85 & 88.17 \\
\cellcolor{c_ours}\hspace{1em}\textbf{+ M\textsuperscript{2}PO} & 88.88 & 88.91 & 88.29 & 90.86 & 87.37 & 89.44 & 88.38 \\
\hline

\cellcolor{c_ours}GemmaX2-28-9B & 88.48 & 88.60 & 87.54 & 90.50 & 86.93 & 89.20 & 88.10 \\
\cellcolor{c_ours}\hspace{1em}+ SFT & 88.75 & 88.38 & 87.90 & 91.23 & 87.28 & 89.55 & 88.17 \\
\cellcolor{c_ours}\hspace{1em}+ CPO & 88.87 & 88.61 & 87.72 & 91.50 & 87.39 & 89.68 & 88.29 \\
\cellcolor{c_ours}\hspace{1em}\textbf{+ M\textsuperscript{2}PO} & \bbcc{89.33} & \bbcc{89.11} & \cc{88.86} & \bbcc{91.93} & 87.67 & 89.79 & 88.61 \\
\bottomrule
\end{tabular}%
}
\caption{\setlength{\fboxsep}{1pt}
Main results on the FLORES-200-test benchmark. Results are reported in \textbf{COMET22} scores. Model names are color-coded by type: \colorbox{c_closed}{Proprietary}, \colorbox{c_open}{Open-source baselines}, and \colorbox{c_ours}{Our experiments}. \textbf{Bold} indicates the best result overall. \colorbox{gray!20}{Colored background} indicates the best result among open-source models.}
\label{tab:flores200_comet22}
\end{table*}

\begin{table*}[t]
\centering
\scriptsize
\resizebox{\textwidth}{!}{%
\setlength{\tabcolsep}{4pt}
\renewcommand{\arraystretch}{1.2}

\begin{tabular}{lcccc}
\toprule
\textbf{Model} & \textbf{AVG} & \textbf{En$\to$Zh} & \textbf{En$\to$De} & \textbf{En$\to$Ja} \\
\midrule
\cellcolor{c_closed}Gemini-2.0-Flash & 97.11 / 86.38 / 84.28 & 96.42 / 81.80 / 83.86 & 98.47 / 92.56 / 82.52 & 96.45 / 84.79 / 86.47 \\
\cellcolor{c_closed}GPT-4o & 97.25 / 86.00 / 84.03 & 96.57 / 80.91 / 83.66 & \bb{98.56} / 92.83 / 82.58 & 96.63 / 84.26 / 85.84 \\
\cellcolor{c_closed}GPT-4o-mini & 97.15 / 84.95 / 83.88 & 96.63 / 79.96 / 83.15 & 98.24 / 90.83 / 82.16 & 96.59 / 84.07 / 86.33 \\
\midrule

\cellcolor{c_open}Aya-expanse-32B & 96.81 / 86.45 / 84.42 & 96.44 / 81.21 / 83.56 & 97.90 / 92.74 / 82.56 & 96.09 / 85.40 / \bbcc{87.14} \\
\cellcolor{c_open}Aya-23-35B & 95.95 / 84.36 / 82.87 & 95.20 / 79.66 / 81.84 & 97.79 / 91.78 / 81.36 & 94.87 / 81.64 / 85.42 \\
\cellcolor{c_open}TowerInstruct-13B & 94.49 / 76.45 / 80.69 & 94.33 / 74.50 / 81.16 & 97.81 / 91.23 / 81.44 & 91.33 / 63.63 / 79.47 \\
\cellcolor{c_open}ALMA-13B-R & 92.68 / 74.84 / 80.30 & 92.93 / 73.38 / 80.53 & 96.77 / 91.46 / 81.25 & 88.35 / 59.68 / 79.13 \\
\midrule

\cellcolor{c_ours}Qwen3-4B-Instruct & 95.61 / 82.31 / 81.74 & 96.09 / 80.60 / 82.98 & 95.76 / 88.29 / 78.41 & 94.97 / 78.04 / 83.82 \\
\cellcolor{c_ours}\hspace{1em}+ SFT & 95.96 / 82.87 / 82.25 & 96.44 / 81.45 / 83.38 & 96.30 / 88.72 / 79.25 & 95.15 / 78.43 / 84.12 \\
\cellcolor{c_ours}\hspace{1em}+ CPO & 96.51 / 83.20 / 82.63 & 96.57 / 81.84 / 83.98 & 97.13 / 89.02 / 79.51 & 95.84 / 78.75 / 84.39 \\
\cellcolor{c_ours}\hspace{1em}\textbf{+ M\textsuperscript{2}PO} & 96.95 / 83.76 / 83.11 & 96.78 / 82.54 / 84.36 & 97.72 / 89.39 / 80.10 & 96.36 / 79.36 / 84.86 \\
\hline

\cellcolor{c_ours}GemmaX2-28-9B & 95.65 / 84.69 / 83.81 & 94.79 / 80.56 / 84.11 & 96.94 / 92.19 / 82.28 & 95.23 / 81.32 / 85.03 \\
\cellcolor{c_ours}\hspace{1em}+ SFT & 96.54 / 86.28 / 84.32 & 96.13 / 81.52 / 84.15 & 97.40 / 92.06 / 82.32 & 96.09 / 85.27 / 86.49 \\
\cellcolor{c_ours}\hspace{1em}+ CPO & 96.72 / 86.71 / 84.51 & 96.29 / 82.09 / 84.35 & 97.59 / 92.31 / 82.42 & 96.28 / 85.73 / 86.77 \\
\cellcolor{c_ours}\hspace{1em}\textbf{+ M\textsuperscript{2}PO} & \bbcc{97.31} / \bbcc{87.34} / \bbcc{84.92} & \bbcc{96.83} / \bbcc{82.94} / \bbcc{84.87} & \cc{98.32} / \bbcc{92.84} / \bbcc{82.89} & \bbcc{96.77} / \bbcc{86.24} / 87.00 \\
\bottomrule
\end{tabular}%
}
\caption{\setlength{\fboxsep}{1pt}
Main results on the WMT24 benchmark (En$\to$X). Results are reported in the format of \textbf{Coverage / XCOMET / COMET22}. Model names are color-coded by type: \colorbox{c_closed}{Proprietary}, \colorbox{c_open}{Open-source baselines}, and \colorbox{c_ours}{Our experiments}. \textbf{Bold} indicates the best result overall. \colorbox{gray!20}{Colored background} indicates the best result among open-source models.}
\label{tab:wmt24_results}
\end{table*}

\end{document}